\documentclass{article}
\usepackage[nonatbib, preprint]{neurips_2020}

\usepackage{url}
\usepackage{latexsym,amssymb}
\usepackage[cmex10]{amsmath}

\usepackage{graphicx}
\usepackage{tikz}
\usepackage{tikz-network}

\usepackage[edges]{forest}
\usetikzlibrary{shadows.blur}

\usetikzlibrary{positioning, decorations.pathmorphing}
\definecolor{nblue}{HTML}{88c0d0}
\definecolor{nwhite}{HTML}{eceff4}
\definecolor{nyellow}{HTML}{ebcb8b}
\definecolor{nred}{HTML}{bf616a}
\definecolor{ngreen}{HTML}{a3be8c}

\usepackage{color, colortbl}
\definecolor{Gray}{gray}{0.92}
\newcolumntype{g}{>{\columncolor{Gray}}c}

\usepackage{caption}
\usepackage{subcaption}
\usepackage{multirow}
\usepackage{booktabs}
\usepackage{bm}
\usepackage{array}
\newcolumntype{C}[1]{>{\centering\let\newline\\\arraybackslash\hspace{0pt}}m{#1}}
\newcolumntype{G}[1]{>{\columncolor{Gray}\centering\let\newline\\\arraybackslash\hspace{0pt}}m{#1}}

\usepackage{pifont}
\newcommand{\xmark}{\ding{55}}%
\newcommand{\cmark}{\ding{51}}%

\newcommand\T{\rule{0pt}{2.9ex}}       
\newcommand\B{\rule[-1.2ex]{0pt}{0pt}} 

\title{A Review of Biomedical Datasets Relating to Drug Discovery: A Knowledge Graph Perspective}

\author{
    Stephen Bonner\textsuperscript{1},\enskip Ian P Barrett\textsuperscript{1},\enskip Cheng Ye\textsuperscript{1},\enskip Rowan Swiers\textsuperscript{1} \\
\rule[10pt]{0pt}{0pt}
\textbf{Ola Engkvist\textsuperscript{2}, Andreas Bender\textsuperscript{4}, Charles Tapley Hoyt \textsuperscript{4}, William Hamilton\textsuperscript{5,6}} \\
\rule[15pt]{0pt}{0pt}
{\normalsize \textsuperscript{1}Data Sciences and Quantitative Biology, Discovery Sciences, R\&D, AstraZeneca, Cambridge, UK}\\
{\normalsize \textsuperscript{2}Molecular AI, Discovery Sciences, R\&D, AstraZeneca, Gothenburg, Sweden}\\
{\normalsize \textsuperscript{3}Centre for Molecular Informatics, Department of Chemistry, University of Cambridge, Cambridge, UK}\\
{\normalsize \textsuperscript{4}Laboratory of Systems Pharmacology, Harvard Medical School, Boston, USA}\\
{\normalsize \textsuperscript{5}School of Computer Science, McGill University, Montreal, Canada} \\
{\, \normalsize \textsuperscript{6}Mila - Quebec AI Institute, Montreal, Canada}\\
\rule[15pt]{0pt}{0pt}
}

\begin{document}
\maketitle 

\begin{abstract}

Drug discovery and development is a complex and costly process. Machine learning approaches are being investigated to help improve the effectiveness and speed of multiple stages of the drug discovery pipeline. Of these, those that use Knowledge Graphs (KG) have promise in many tasks, including drug repurposing, drug toxicity prediction and target gene-disease prioritisation. In a drug discovery KG, crucial elements including genes, diseases and drugs are represented as entities, whilst relationships between them indicate an interaction.

However, to construct high-quality KGs, suitable data is required. In this review, we detail publicly available sources suitable for use in constructing drug discovery focused KGs. We aim to help guide machine learning and KG practitioners who are interested in applying new techniques to the drug discovery field, but who may be unfamiliar with the relevant data sources. The datasets are selected via strict criteria, categorised according to the primary type of information contained within and are considered based upon what information could be extracted to build a KG. We then present a comparative analysis of existing public drug discovery KGs and a evaluation of selected motivating case studies from the literature. Additionally, we raise numerous and unique challenges and issues associated with the domain and its datasets, whilst also highlighting key future research directions. We hope this review will motivate KGs use in solving key and emerging questions in the drug discovery domain.

\end{abstract}

\section{Introduction}
\label{sec:intro}

Discovering new drugs is a complex task, requiring knowledge from numerous biological and chemical domains, as well as understanding in various subtasks. For example, drugs are primarily developed in response to some disease or condition negatively affecting patients. This implicitly requires that the mechanisms in the body by which the disease is caused are sufficiently understood so that a drug can be designed to treat it – a process known as target discovery~\cite{lindsay2003target}. However, due to the complexities involved, the process of developing a new drug and bringing it to market is expensive and has a high chance of failure~\cite{morgan2018impact}.

Hence, researchers are striving to increase the probability of success for the drug discovery process. Graphs\footnote{Also commonly known as networks within the biological domain. In this review we use the term graph interchangeably with network and without loss of generality.} have long been used in the life sciences as they are well suited to the complex interconnected systems often studied in the domain~\cite{barabasi2004network}. Homogeneous graphs have, for example, been used extensively to study protein-protein interaction networks~\cite{neyshabur2013netal}, where each vertex represents a protein, and edges capture interactions between them. 

Recently Knowledge Graphs (KGs) have begun to be utilised to model various aspects of the drug discovery domain. KGs are heterogeneous data representations and build upon the linked open data and semantic web principles~\cite{jupp2014ebi}. In a KG, both the vertices and edges can be of multiple different types, allowing for more complex and nuanced relationships to be captured~\cite{hogan2021knowledge}. In the context of drug discovery, the entities represent key elements such as genes, disease or drugs -- with the edge types capturing different categories of interaction between them. As an example of where having distinct edge types could be crucial, an edge between a drug and disease entity could indicate that the drug has been clinically successful in treating the disease. Conversely, an edge between the same two entities could mean the drug was assessed but ultimately proved unsuccessful. This crucial distinction in the precise meaning of the relationship between the two entities would not truly be captured in the simple binary option offered by homogeneous graphs, whereas, a KG representation would preserve this important difference and enable that knowledge to be used to inform better predictions. As a topical concrete application, KGs have been utilised to address various tasks in helping to combat the COVID-19 pandemic~\cite{bettencourt2020exploring, cernile2020network, domingo2020covid, drkg2020, reese2020kg, wise2020covid}. Additionally, considering the domain as a KG has the potential to enable recent advances in graph-specific machine learning to be exploited~\cite{gaudelet2020utilising}.

However, constructing a suitable and informative KG requires that the correct primary data is captured in the process. An interesting aspect of the drug discovery domain, and perhaps in contrast to others, is that there is a wealth of well curated, publicly available data sources, many of which can be represented as, or used to construct, KGs~\cite{rigden202027th}. Many of these are maintained by government and international level agencies and are regularly updated with new results~\cite{rigden202027th}. Indeed, one could argue that there is sometimes too much data available, rather than too little, and researchers working in drug discovery must instead consider other issues when looking to use these data resources with graph analytics. Such issues include assessing how reliable the underlying information is, how best to integrate disparate and heterogeneous resources, how to deal with the uncertainty inherent in the domain, how best to translate key drug discovery objectives into machine learning training objectives, and how to model and express data that is often quantitative and contextual in nature. Despite these complications, an increasing level of interest  suggests that KGs could play a crucial role in enabling machine learning approaches for drug discovery~\cite{gaudelet2020utilising, himmelstein2017systematic, zitnik2018modeling}.

We present a review of the publicly available data sources for drug discovery, detailing how they could be utilised in a KG setting and analysing existing pre-constructed graphs. To the best of our knowledge, this is the first time these resources have been compared and evaluated in the literature. The primary contributions of this review are as follows:

\begin{itemize}
	\item We present an introduction to the drug discovery domain for KG and machine learning practitioners, whilst detailing the numerous unique research challenges it poses.
	\item We review key data sources within drug discovery, present a taxonomy based on their primary biomedical area and consider how amenable they are for use in KGs by detailing what type of information could be extracted from them (relational versus entity features).
	\item We perform a comparative analysis of existing public drug discovery KGs based on their underlying data sources and graph composition decisions.
	\item We detail motivating case studies of KG use within drug discovery.
	\item We outline the key directions for future research and open problems within the domain.
\end{itemize}

Our hope is that this review will serve as motivation for researchers and enable greater, easier and more effective use of KGs in drug discovery by signposting key resources in the field and highlighting some of the primary challenges. We aim to help foster a multi-disciplinary and collaborative outlook that we believe will be critical in considering graph composition and construction in concert with analytical approaches and clarity of purpose. We think the review will also be useful for researchers in the drug discovery domain who are interested in the potential insights to be gained by applying KG methodologies.

An open-source collection of the resources detailed in this review has also been released.\footnote{\url{https://github.com/AstraZeneca/awesome-drug-discovery-knowledge-graphs}}

\subsection{Dataset Selection Criteria}

For the purpose of this review, we use the following criteria when choosing datasets for inclusion:

\begin{itemize}
	\item \emph{Publicly Accessible - } The dataset should be available for use within the public domain. Whilst many high quality commercial datasets exist, we choose to focus on only those datasets which are publicly accessible to some degree.
	\item \emph{High Quality - } The dataset should contain information of the highest quality. This will primarily be assessed through its popularity within the drug discovery literature.
	\item \emph{Actively Maintained - } The dataset should still be actively maintained and updated.
\end{itemize}

\subsection{Review Organisation} The remainder of the survey is structured as follows: in Section \ref{sec:background} we introduce the required background knowledge, Section \ref{sec:competing_studies} details existing work, Section \ref{sec:ontologies} introduces the relevant ontologies, in Section \ref{sec:datasets} the primary datasets are reviewed, in Section \ref{sec:exisiting_kgs} existing drug discovery knowledge graphs are detailed, Section \ref{sec:case_study} presents some application case studies, Section \ref{sec:future} details future challenges for the field and the final conclusions are presented in Section \ref{sec:conclusion}.
\section{Background}
\label{sec:background}

In this section, we introduce key background concepts including KGs and the field of drug discovery.

\subsection{An Introduction To Drug Discovery and Development}
\label{ssec:drug_discovery_intro}

Drug discovery and development is a complex and highly multi-disciplinary process~\cite{terstappen2001silico} and is driven by the need to address a disease or other medical condition affecting patients for which no suitable treatment is currently in production, or where current treatments are insufficient~\cite{hughes2011principles}. Whilst a full review of the area is beyond the scope of our own review (interested readers are referred to relevant reviews~\cite{cook2014lessons, morgan2018impact, wagner2018dynamic}) here we present a high-level overview of key concepts. This section will make use of many of the biological terms and concepts defined in Table \ref{tab:definitions}. 

Drug discovery involves searching for causally implicated molecular functions, biological and physiological processes underlying disease, and designing drugs that can modify, halt or revert them. There are currently three main routes to drug discovery – selecting a molecular target(s) against which to design a drug (targeted drug discovery), designing a high-throughput experiment to act as a surrogate for a disease process and then screening molecules to find ones that affect the outcome (phenotypic drug discovery), or using an existing drug developed for another disease (drug repositioning). In targeted drug discovery, once a drug target has been identified, the process of finding suitable drug compounds can begin via an iterative drug screening process. Selected possible candidate drugs are then tested through a series of experiments in preclinical models (both \emph{in vitro} – study in cells or artificial systems outside the body, and \emph{in vivo} – study in a whole organism), and then clinical trials (drug development) to measure efficacy (beneficial modification of disease process) and toxicity (undesirable biological effects). 

\begin{table}[!ht]
    \centering
	\renewcommand{\arraystretch}{1.5}
    \resizebox{\textwidth}{!}{
    \begin{tabular}{>{\centering\arraybackslash} m{2.5cm} m{16cm} } 
    \toprule
    \textbf{Term} & \textbf{Definition} \T\B \\
	\midrule \midrule
	
	\emph{Cells} & Important for preclinical research and data generation, e.g. studying drug responses, gene and protein expression, morphological responses via imaging.\\ \cmidrule{2-2}

    \emph{Genes} & Functional units of DNA, encoding RNA, and ultimately proteins. Variants of a gene’s DNA sequence may be associated with disease(s). \\
    \cmidrule{2-2}
	\emph{Transcripts} & Gene sequences are transcribed into an intermediate molecule called RNA, which is in turn “read” to produce protein molecules. \\
	\cmidrule{2-2}
	\emph{Proteins} & Key functional units of a cell and that can play structural or signalling roles, or catalyse reactions, and interact with other proteins. \\
	\cmidrule{2-2}
	\emph{Biological Processes \& Pathways} & Molecules and cells function together to perform biological processes which can be conceptualised at different scales, from intracellular (e.g. signalling transmission via “pathways” between molecules) to intercellular processes, and ultimately physiological processes at the tissue and body scale.\\
	\cmidrule{2-2}
	\emph{Diseases} & A condition resulting from aberrant biological/physiological processes. Different diseases may share symptoms and underlying aberrant processes, and can often be categorised into subtypes based on clinical and/or molecular features. \\
	\cmidrule{2-2}
	\emph{Targets} & A drug target is a molecule(s) whose modulation (by a drug) we hypothesise will alter the course of the disease. \\
	\cmidrule{2-2}
	\emph{Compounds} & Small molecules generated and studied as part of drug discovery are sometimes termed “compounds”, with an accompanying chemical structure representation. \\
	\cmidrule{2-2}
	\emph{In Vitro} & Studies and experiments taking place outside the body, either in cells or in cell-free, highly defined systems. \\
	\cmidrule{2-2}
	\emph{In Vivo} & Studies and experiments taking place in a physiological context (e.g. animal or human study). \\

    \bottomrule
    \end{tabular}}
    \caption{Definitions of key terms used within the scope of drug discovery.}
	\label{tab:definitions}
	\vskip -15pt
\end{table}

The active molecules in drugs have most often been small chemicals (sometimes termed compounds) and antibodies (a type of protein)~\cite{blanco2020new}, but can also be other types such as peptides, nucleotides, macromolecules, or polymers. Various newer types of drugs, often collectively termed \emph{“drug modalities”} are also being explored~\cite{blanco2020new}. These different types of drugs have particular advantages and disadvantages, but together open up a wider set of potential drug targets compared to historical approaches alone.

Biomedical science has researched the various processes highlighted in this section at different scales, using technologies that probe the abundance and sequence variation in DNA, RNA and proteins, and for studying specific biological functions via experimentation. For example, studies in genetic variation associated with disease are used to provide support to hypotheses for new drug targets~\cite{nelson2015support, king2019drug}. Databases have been constructed to collate and disseminate such data and information~\cite{rigden202027th} (detailed in Section \ref{sec:datasets}). Ontologies have also been developed to model relevant concepts such as disease and are discussed more in Section \ref{sec:ontologies}.

\subsubsection{Subtasks Within Drug Discovery}

The field is increasingly looking towards computational~\cite{terstappen2001silico} and machine learning approaches to help in various tasks within the drug discovery process~\cite{vamathevan2019applications}. It can be helpful to consider partitioning the drug discovery process up into smaller subtasks, some of the most common being:

\begin{itemize}
	\item \emph{Drug Repositioning} - Which drugs previously tested in clinical trials could be ascribed new indications?

	\item \emph{Disease Target Identification} - Which molecular entities (genes and proteins) are implicated in causing or maintaining a disease? Also known as Target Identification and Gene-Disease Prioritisation.
	
	\item \emph{Drug Target Interaction} - Given a drug with unknown interactions, what proteins may it interact with in a cell? Also known as Target Binding and Target Activity.
	
	\item \emph{Drug Combinations} - What are the beneficial, or toxicity consequences of more than one drug being present and interacting with the biological system?
	
	\item \emph{Drug Toxicity Predictions} – What toxicities may be produced by a drug, and in turn which of those are elicited by modulating the intended target of the drug, and which are from other properties of the drug? Also known as Toxicity Prediction. 
\end{itemize}

\subsection{Knowledge Graphs}
\label{ssec:kg}

There is currently not a strict and commonly agreed upon definition of a KG in the literature~\cite{hogan2021knowledge}. Whilst we do not aim to give a definitive definition here, we instead define KGs as they will be used through the remainder of this work. We first start by defining homogeneous graphs, before expanding the definition for heterogeneous graphs.

\begin{figure}[!ht]
	\centering
	\resizebox{0.7\textwidth}{!}{
	\begin{subfigure}[b]{0.49\textwidth}
		\centering
	
	    \begin{tikzpicture}
			\Vertex[IdAsLabel, Math, size=1.]{v1}
			\Vertex[IdAsLabel, Math, size=1., x=3]{v2}
			\Vertex[IdAsLabel, Math, size=1., x=2.5,y=-2]{v3}
			\Vertex[IdAsLabel, Math, size=1., x=-1.7,y=1.5]{v4}
			\Vertex[IdAsLabel, Math, size=1., x=-1.9,y=-1.4]{v5}
			\Vertex[IdAsLabel, Math, size=1., x=1,y=-3.2]{v6}
			\Vertex[IdAsLabel, Math, size=1., x=.9,y=2.6]{v7}

			\Edge[Direct, lw=2pt](v1)(v2)
			\Edge[Direct, lw=2pt](v1)(v3)
			\Edge[Direct ,lw=2pt](v1)(v4)
	
			\Edge[Direct, lw=2pt](v5)(v4)
	
			\Edge[lw=2pt](v2)(v7)
	
			\Edge[lw=2pt](v3)(v6)
			\Edge[lw=2pt](v1)(v6)
			\Edge[lw=2pt](v5)(v6)
			
		\end{tikzpicture}

	\caption{A Homogeneous Graph.}\label{fig:homo}
	\end{subfigure}
	\hfill
	\begin{subfigure}[b]{0.49\textwidth}
		\centering		
		\begin{tikzpicture}
			\Vertex[IdAsLabel, Math, size=1., RGB, color={190,174,212}]{v1_1}
			\Vertex[IdAsLabel, Math, size=1., x=3, RGB, color={163,190,140}]{v2_1}
			\Vertex[IdAsLabel, Math, size=1., x=2.5,y=-2]{v3_1}
			\Vertex[IdAsLabel, Math, size=1., x=-1.7,y=1.5]{v3_2}
			\Vertex[IdAsLabel, Math, size=1., x=-1.9,y=-1.4, RGB ,color={163,190,140}]{v2_2}
			\Vertex[IdAsLabel, Math, size=1., x=1,y=-3.2]{v3_3}
			\Vertex[IdAsLabel, Math, size=1., x=.9,y=2.6, RGB, color={190,174,212}]{v1_2}

			\Edge[Direct, lw=2pt, bend=-10, label=e1, RGB, color={191, 97, 106}](v1_1)(v2_1)
			\Edge[Direct, lw=2pt, bend=-10, label=e2, RGB, color={235, 203, 139}](v2_1)(v1_1)
			\Edge[Direct, lw=2pt, label=e2, RGB, color={235, 203, 139}](v1_1)(v3_1)
			\Edge[Direct ,lw=2pt, label=e1, RGB, color={191, 97, 106}](v1_1)(v3_2)
	
			\Edge[Direct, lw=2pt, label=e1, RGB, color={191, 97, 106}](v2_2)(v3_2)
	
			\Edge[lw=2pt, RGB, label=e3, color={208, 135, 112}](v2_1)(v1_2)
	
			\Edge[lw=2pt, RGB, label=e3, color={208, 135, 112}](v3_1)(v3_3)
			\Edge[Direct, lw=2pt, label=e1, RGB, color={191, 97, 106}](v6)(v1_1)
			\Edge[Direct, lw=2pt, label=e2, bend=-10, RGB, color={235, 203, 139}](v3_3)(v2_2)
			\Edge[Direct, lw=2pt, label=e2, bend=10, RGB, color={235, 203, 139}](v3_3)(v2_2)

		\end{tikzpicture}

	\caption{A Heterogeneous Graph.}\label{fig:het}
	\end{subfigure}}

	\caption{A Homogeneous and Heterogeneous Graph.}
	\vskip -10pt
	\label{fig:homo-v-het}
\end{figure}
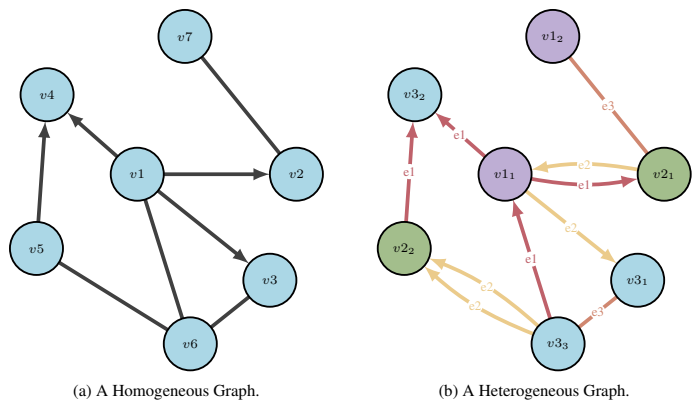

A homogeneous graph can be defined as \(G = (V,E)\) where \(V\) is a set of vertices and \(E\) is a set of edges. The elements in \(E\) are pairs \( (u,v) \) of unique vertices \(u,v \in V\). An example graph is illustrated in Figure \ref{fig:homo} which demonstrates that homogeneous graphs can contain a mix of directed and undirected edges. It is possible for these graphs to have a set of features associated with the vertices, typically represented as a matrix \(X\in \mathbb{R}^{|V| \times f} \), where \(f\) is the number of features for a certain vertex.

Heterogeneous graphs, or KGs as they are termed, are graphs which contain distinct different types of both vertices and edges, which can be defined as  \(G=(V,E, R, \Psi )\) ~\cite{zhang2019heterogeneous}. Such graphs now have a set of relations \(R\), and each edge is now defined by its relation type \(r \in R\) -- meaning that edges are now represented as triplet values \((u,r,v) \in E\)~\cite{lee2020heterogeneous}. The vertices in knowledge graphs are often known as entities, with the first entity in the triple called the head entity, connected via a relation to the tail entity. In a drug discovery context, multiple relations are crucial as an edge could indicate whether a drug up or down regulates a certain gene for example. Two vertices can also now be linked by more than one edge type, or even multiples of the same type. Again this is important in the drug discovery domain, as multiple edges of the same relation can indicate evidence from multiple sources. Additionally, each vertex in a heterogeneous graph also belongs to a certain type from the set \(\Psi\), meaning that our original set of vertices can be divided into subsets \(V_i \subset V \), where \(i \in \Psi\) and \(V_i \cap V_j = \emptyset, \forall i \in \Psi \ne j \in \Psi \)~\cite{hamilton2020graph}. Given the drug discovery focus, these types could indicate if a vertex represents a gene, protein or drug. Further, these vertex types can limit the type of relations placed between them, \((u,r_1,v) \in E \to u \in V_i,v \in V_j\) where \(i,j \in \Psi \)~\cite{hamilton2020graph}. An edge relation type of \emph{`expressed-as'} makes sense between genes and proteins but not genes and drugs for example.

A heterogeneous graph is presented in Figure \ref{fig:het} and contains some key differences with its homogeneous counterpart: there are three types of vertex (v1, v2 and v3) and these are linked through a mix of directed and undirected edges of three relation types (e1, e2 and e3).

\subsection{Knowledge Graph Use in Drug Discovery: Research Challenges}
\label{ssec:graph-ml-kg}

\begin{figure}[!ht]
	\centering
	\includegraphics[width=0.5\textwidth]{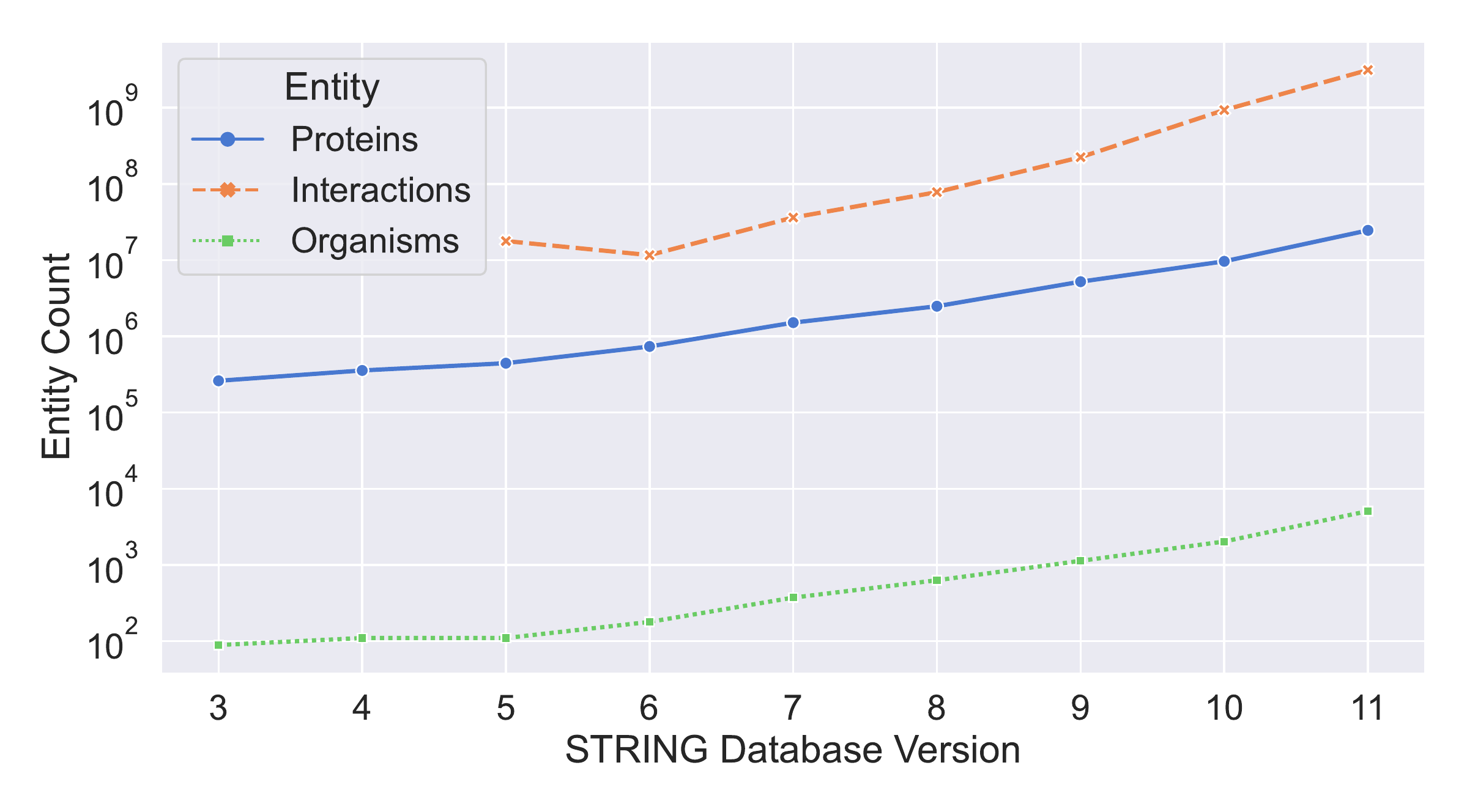}
	\caption{The evolution of the STRING database over major versions showing the increase in Organisms, Interactions and Proteins.}
	\label{fig:STRING-db}
	\vskip -10pt
\end{figure}

The study of KGs in the biomedical sciences and particularly drug discovery brings challenges and opportunities. Opportunities because biomedical information inherently contains many relationships which can be exploited for new knowledge. Unfortunately there are many challenges that arise when constructing a KG suitable for drug discovery tasks, particularly when combined with machine learning. Some of the most interesting research challenges are detailed below:

\begin{itemize}
	\item \emph{Graph Composition} - Strategies are needed to define how to convert data into information for modelling in a graph (e.g., instantiating a node or edge versus a feature on those entities), and what scale and composition of graph(s) may be optimal for a given task. Is a single large graph best, or should task specific graphs be constructed? In addition, which type of analytical approach to use - reasoning-based, network/graph theoretical, machine learning, or hybrid approaches.
	
	\item \emph{Heterogeneous \& Uncertain} - In biomedical graphs, the data types are heterogeneous and have differing levels of confidence (e.g. well characterised and curated findings versus NLP-derived assertions), and much of the data will be dependent on numerous factors both time and the dose of drug used as well as the genetic background in the study. Overall, this means edges are much less certain, and thus less trustworthy, than in other domains. 
	
	\item \emph{Evolving Data} - The underlying data sources integrated and used in suitable KGs are also often changing over time as the field develops, requiring attention to versioning and other reproducible research practices. As an example of this, the evolution of the frequently used STRING dataset is demonstrated in Figure \ref{fig:STRING-db}.\footnote{This data has been collected from \url{https://string-db.org/cgi/access.pl?sessionId=dbw44gRWU7Xo&footer_active_subpage=archive}}
	
	\item \emph{Bias} - There are various biases evident in different data sources, for example negative data remain under-represented in some sources, including the primary scientific literature, and some areas have been studied more than others, introducing ascertainment bias in the graphs~\cite{oprea2018unexplored}. 
	
	\item \emph{Fair Evaluation} - Several works have shown promise in applying machine learning techniques on a KG of drug discovery data. However, ensuring a fair data split is used for evaluation is perhaps more complicated than other domains, as it is easy for biologically or chemically meaningful data to leak across train/test splits. Thus, care should be taken to construct more meaningful data splits, as well as considering if replicated knowledge has been incorporated in the graph and could potentially leak across from the train/test split. For example, proteochemometrics approaches often employ a clustering-based splitting of chemicals to reduce leakage of similar chemicals between the training and testing sets~\cite{Lopez-DelRio2019}.

	\item \emph{Meaningful Evaluation} - While most practical applications of link prediction only focus on a single relation type (e.g., chemical modulates protein), metrics are often reported as an aggregate over all relation types. Because bias could be introduced by a large number of relations of other types either scoring much better or worse than the target relation type on average, leading to an inaccurate evaluation, metrics should be reported broken down by relation type.

	\item \emph{Beware of Metrics} - Because common metrics used in link prediction tasks like mean rank (MR), mean reciprocal rank (MRR), and Hits at \textit{K} are not comparable on results from KGs of different sizes, alternative metrics like the adjusted mean rank (AMR) should be employed~\cite{berrendorf2020ambiguity}. Different implementations of link prediction evaluation calculate metrics very differently and caution should be observed when comparing results from different packages. Further, link prediction models built on biological KGs often influence real-world experimentation, so discussion on evaluation metrics should be considered with respect to how it can help achieve real-world goals.
\end{itemize}

Ultimately we feel there is now an interesting opportunity to experiment at the intersection of various research fields spanning graph theoretic and other network analysis approaches for molecule networks~\cite{barabasi2011network, choobdar2019assessment}, machine learning approaches~\cite{zitnik2018modeling}, and quantitative systems pharmacology~\cite{sorger2011quantitative}.

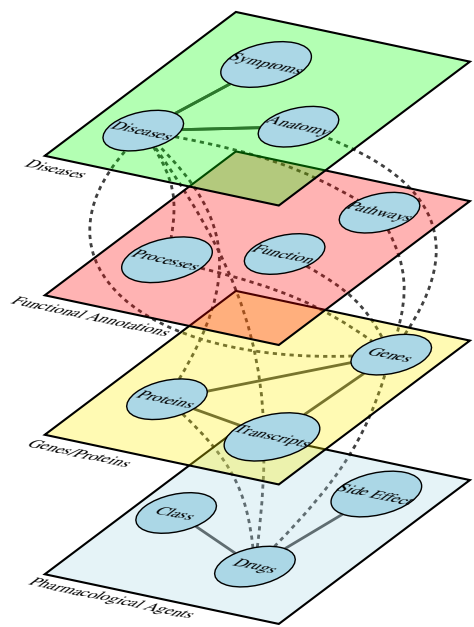
\begin{figure}[!ht]
	\centering
	\resizebox{.45\textwidth}{!}{
	\begin{tikzpicture}[multilayer=3d]
  
		\SetLayerDistance{-3.5}
		\Plane[x=-.5,y=-.5, width=6, height=6, layer=4]
		\Plane[x=-.5,y=-.5, width=6, height=6, layer=3, color=yellow]
		\Plane[x=-.5,y=-.5, width=6, height=6, layer=2, color=red]
		\Plane[x=-.5,y=-.5, width=6, height=6, layer=1, color=green]

		\Text[rotation=360, x=0.1, y=-0.85, layer=1, fontsize=\large]{Diseases}

		\Vertex[IdAsLabel, size=1.6, x=0.8, y=1., layer=1, fontsize=\large]{Diseases}
		\Vertex[IdAsLabel, size=1.8, x=1.5, y=4, layer=1, fontsize=\large]{Symptoms}
		\Vertex[IdAsLabel, size=1.6, x=3.8, y=2.2, layer=1, fontsize=\large]{Anatomy}

		\Edge[lw=2pt, bend=0](Diseases)(Symptoms)
		\Edge[lw=2pt, bend=0](Diseases)(Anatomy)

		\Vertex[IdAsLabel, size=1.8, x=1, y=1.5, layer=2, fontsize=\large]{Processes}
		\Vertex[IdAsLabel, size=1.6, x=4, y=4.5, layer=2, fontsize=\large]{Pathways}
		\Vertex[IdAsLabel, size=1.6, x=3.2, y=2.5, layer=2, fontsize=\large]{Function}

		\Edge[lw=2pt, bend=0](Diseases)(Symptoms)
		\Edge[lw=2pt, bend=0](Diseases)(Anatomy)

		\Text[rotation=360, x=1.0, y=-0.85, layer=2, fontsize=\large]{Functional Annotations}

		\Vertex[IdAsLabel, size=1.6, x=1, y=1.5, layer=3, fontsize=\large]{Proteins}
		\Vertex[IdAsLabel, size=1.6, x=4.3, y=4.5, layer=3, fontsize=\large]{Genes}
		\Vertex[IdAsLabel, size=1.9, x=4.1, y=1.0, layer=3, fontsize=\large]{Transcripts}

		\Edge[lw=2pt, bend=0](Proteins)(Genes)
		\Edge[lw=2pt, bend=0](Proteins)(Transcripts)
		\Edge[lw=2pt, bend=0](Transcripts)(Genes)

	  	\Text[rotation=360, x=.7, y=-0.85, layer=3, fontsize=\large]{Genes/Proteins}

		\Vertex[IdAsLabel, size=1.6, x=3.5, y=1.2, layer=4, fontsize=\large]{Drugs}
		\Vertex[IdAsLabel, size=1.8, x=4, y=4.4, layer=4, fontsize=\large]{Side Effect}
		\Vertex[IdAsLabel, size=1.6, x=0.5, y=2.4, layer=4, fontsize=\large]{Class}

		\Edge[lw=2pt, bend=0](Drugs)(Side Effect)
		\Edge[lw=2pt, bend=0](Drugs)(Class)

		\Text[rotation=360, x=1.5, y=-0.85, layer=4, fontsize=\large]{Pharmacological Agents}

		\Edge[lw=2pt, bend=10,style=dashed](Processes)(Genes)
		\Edge[lw=2pt, bend=10,style=dashed](Function)(Genes)
		\Edge[lw=2pt, bend=10,style=dashed](Pathways)(Genes)

		\Edge[lw=2pt, bend=10,style=dashed](Proteins)(Drugs)
		\Edge[lw=2pt, bend=-10,style=dashed](Drugs)(Genes)

		\Edge[lw=2pt, bend=10, style=dashed](Diseases)(Processes)
		\Edge[lw=2pt, bend=10, style=dashed](Diseases)(Pathways)

		\Edge[lw=2pt, bend=50, style=dashed](Genes)(Diseases)
		\Edge[lw=2pt, bend=-30, style=dashed](Genes)(Anatomy)
		\Edge[lw=2pt, bend=-20, style=dashed](Proteins)(Diseases)

		\Edge[lw=2pt, bend=-10, style=dashed](Drugs)(Diseases)
		
	\end{tikzpicture}
	}
	\caption{A simplified hierarchical view of a drug discovery knowledge graph schema.}
	\vskip -15pt
\end{figure}
\section{Prior Work}
\label{sec:competing_studies}

The area of drug repurposing has been addressed in several reviews~\cite{tanoli2020exploration, luo2020biomedical, zhu2020knowledge, masoudi2020drug}. Recent worked has detailed over 100 relevant drug repurposing databases, as well as appropriate methods~\cite{tanoli2020exploration}. In~\cite{luo2020biomedical}, a review of repurposing from the view of machine learning is been presented, covering methods and over 20 datasets. KG specific approaches for repurposing have been reviewed, with the authors detailing suitable datasets then choosing 6 to form the KG used in their experimental evaluation~\cite{zhu2020knowledge}. In~\cite{masoudi2020drug}, the authors review and then partition the available drug databases into four categories based upon the type of information contained within: raw data, target-based, area specific and drug design.

The area of drug-target interaction has been reviewed~\cite{bagherian2020machine, chen2018machine}, both focusing upon the various methods for predicting interactions, however potential data sources are also presented. Conversely, machine learning based approaches for predicting drug-drug interaction have been detailed, with comparative evaluations conducted~\cite{celebi2019evaluation}. The authors construct a drug-drug interaction KG from a subset of Bio2RDF~\cite{belleau2008bio2rdf}. A review of 13 drug related databases has been presented~\cite{zhu2019drug}, covering a broad range of databases detailing drugs and drug-target interactions.

One study reviews both datasets and approaches for biological KG embeddings~\cite{mohamed2020biological}. Although the review focuses upon the evaluation of different methodologies, 16 relevant databases are also discussed. However, as the work is experimentally driven, only a limited dataset discussion is undertaken. A different survey of the wider biomedical area and KG use within it has been presented~\cite{callahan2020knowledge}. Finally, a recent study presents a detailed overview of the application of graph-based machine learning in drug discovery~\cite{gaudelet2020utilising}. The review is wide-ranging but makes no mention of suitable public datasets. We do however feel that it strongly complements our own review and serves as a method-focused counterpart to our dataset overview.    

One issue with existing reviews is that they are focused upon a specific subtask, thus they are not giving a clear view of the overall drug discovery landscape. Another trend in the current reviews is for their primary focus to be upon the experimental evaluation of methodologies, with datasets given comparatively less attention. Additionally, most reviews are considering the resources solely as databases, rather than focusing upon how they could form part of a KG. Finally, many of the reviews have been written from a biological point of view, which may make them less accessible for machine learning practitioners who may be new to the domain, but who are interested in experimenting with relevant datasets.
\section{Biomedical Ontologies}
\label{sec:ontologies}

An ontology is a set of controlled terms that defines and categorises objects in a specific subject area. Modern biomedical ontologies are usually human constructed representations of a domain, capturing key entities and relationships and distilling the knowledge into a concise machine readable format~\cite{en2001ontologies}. There is a need for consistency when discussing concepts like diseases and gene functions which can be interpreted in multiple ways. 

\subsection{Ontology Representations}

Most biomedical ontologies are expressed in a knowledge representation language such as the OBO language created by the Open Biological and Biomedical Ontologies Foundry (OBO), Resource Description Framework Schema (RDFS) or the Web Ontology Language (OWL)~\cite{antoniou2004semantic}. OBO is a biologically oriented ontology and is expressive enough to define the required terms, relationships and properties of an ontology, whereas, both OWL and RDFS originate from the semantic web world but are used for creating biomedical ontologies.

\subsection{Ontology Matching and Merging} 

There has been a proliferation of ontologies relating to biomedical data, which can cause issues. For example, if database A labels diseases using ontology X and database B labels diseases using ontology Y, it can be hard to know the relation of two different disease entities. Some resources exist to match together ontological terms; e.g. OXO~\cite{jupp2017oxo} and DODO~\cite{franccois2020dictionary}. However the mappings provided by these resources are different and between any two distinct ontologies, there is no guarantee of a direct mapping or indeed any at all between their ontological terms.

Merging two different ontologies into one is an active area of research. However just mapping terms between ontologies can create logical inconsistencies in the newly created ontology. Therefore merging ontologies often involves manual intervention and is a time consuming and error prone process. The Open Biological and Biomedical Ontologies Foundry was set up to provide rules and advice for ontologies to make them easier to merge and match~\cite{tirmizi2011mapping}.

\subsection{Ontology Overviews}

This section details the major ontologies which are relevant for use in drug discovery tasks, which are detailed in Table \ref{tab:ontology}. Note that a full review is beyond the scope of this work and interested readers are referred to a dedicated review~\cite{rubin2008biomedical}.

\begin{table}[ht!]   
    \centering
    \resizebox{0.8\textwidth}{!}{
    \begin{tabular}{p{0.38\textwidth} C{0.18\textwidth}  C{0.15\textwidth}  C{0.15\textwidth} C{0.1\textwidth}  C{0.15\textwidth} C{0.2\textwidth}} 
    \toprule
    \textbf{Ontology Name}  & \textbf{Primary Domain} & \textbf{Classes} & \textbf{Number of Properties} & \textbf{Max Depth} & \textbf{License}\T\B \\
    \midrule \midrule
    
    Monarch Disease Ontology (MonDO)  & Diseases & 24K & 25 & 16 & Creative Commons\T\B \\

    Medical Subject Headings (MeSH)  & Medical Terms & 300K & 38 & 15 & Custom\T\B \\

    Human Phentoype Ontology (HPO)  & Disease Phenotype & 19K  & 0 & 16 & Custom\T\B \\

    Disease Ontology (DO)  & Diseases & 19K  & 4 & 33 & Creative Commons\T\B \\

    Drug Target Ontology (DTO)  & Drug Targets & 10K & 43 & 11 & Creative Commons\T\B \\

    Gene Ontology (GO)  & Genes & 44K  & 11 & - & Creative Commons\T\B \\

    Experimental Factor Ontology (EFO) & Integrator & 28K  & 66 & 20 & Apache 2.0\T\B \\

    \bottomrule
    \end{tabular}}
    \caption{An overview of Ontologies suitable for use in drug discovery.}
    \label{tab:ontology}
    \vskip -15pt
\end{table}

\subsubsection{Disease Ontologies} Due to the complexities associated with properly defining, categorising and linking diseases, a large number of ontologies have been developed. Prominent examples include the Medical Subject Headings (MeSH)~\cite{lipscomb2000medical}, Human Disease Ontology (DO)~\cite{schriml2019human}, Human Phenotype Ontology (HPO)~\cite{robinson2008human} and Monarch Disease Ontology (MonDO)~\cite{mungall2017monarch}. These typically differ in their intended use-case, for example DO was designed to help in the linking of different datasets, MeSH was created to aid in the indexing of MEDLINE/PubMed articles, HPO describes the phenotypes (the observable traits) of disease, MonDO was designed to harmonise disease definitions between other ontologies. 

\subsubsection{Gene Related Ontologies} The function of genes and associated products is also frequently captured in ontologies, with common ones used in the construction of biomedical KGs including Gene Ontology (GO)~\cite{gene2004gene} and the Drug Target Ontology (DTO)~\cite{lin2017drug}. GO focuses on defining gene activity on the molecular level, linking genes to locations in the body where its function is performed and establishing links between genes and biological processes. In contrast, DTO focuses on linking gene products in relation to drug discovery considerations such as druggability.

\subsubsection{Integrator Ontologies} The Experimental Factor Ontology (EFO) was created by the European Bioinformatics Institute to provide a systematic description of experimental variables available in its databases including disease, anatomy, cell type, cell lines, chemical compounds and assay~\cite{malone2010modeling}. The Open Targets Platform (Section \ref{ssec:intergreated-datasets}) uses EFO to provide the description, phenotypes, cross-references, synonyms, ontology and classification for annotating disease entities.
\section{Primary Domain-Specific Datasets}
\label{sec:datasets}

The drug discovery domain has a wealth of public datasets, many of which have dedicated teams updating them. These include national or international level bodies, for example the US based National Center for Biotechnology Information (NCBI) or the European Bioinformatics Institute (EBI). Additionally the pan-European ELIXIR body, an organisation dedicated to detailing best practices for biomedical data and enabling stable funding, maintains a list of core data resources~\cite{durinx2016identifying}.

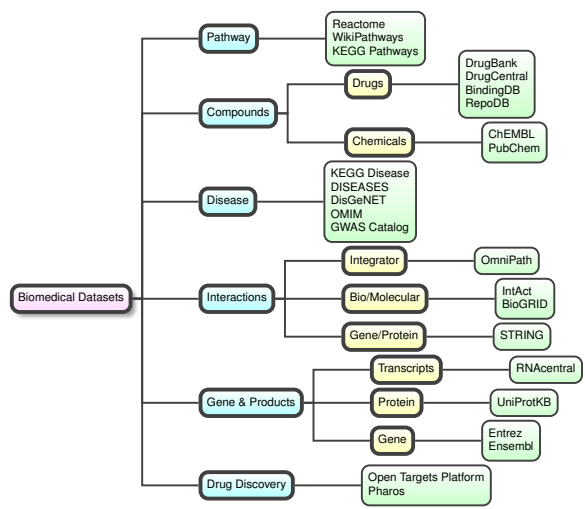
\begin{figure*}
    \centering
    \resizebox{0.55\textwidth}{!}{
    \begin{forest}
        forked edges,
        for tree={font=\tiny\sffamily, rounded corners, top color=gray!1, bottom color=gray!30, edge+={darkgray, line width=1pt}, draw=darkgray, align=left, anchor=west, grow=east, l sep=12mm, s sep=0.5mm},
        before packing={where n children=3{calign child=2, calign=child edge}{}},
        before typesetting nodes={where content={}{coordinate}{}},
         where level<=1{line width=2pt}{line width=1pt},
        [Biomedical Datasets, blur shadow, bottom color=magenta!15
          [Drug Discovery, bottom color=cyan!30
           [Open Targets Platform\\Pharos, bottom color=green!20]
          ]
          [Gene \& Products, bottom color=cyan!30
           [Gene, line width=2pt, bottom color=yellow!30 [Entrez\\Ensembl, bottom color=green!20]]
           [Protein, line width=2pt, bottom color=yellow!30 [UniProtKB, bottom color=green!20]]
           [Transcripts, line width=2pt, bottom color=yellow!30 [RNAcentral, bottom color=green!20]]
          ]
          [Interactions, bottom color=cyan!30, calign with current
           [Gene/Protein, line width=2pt, bottom color=yellow!30 [STRING, bottom color=green!20]]
           [Bio/Molecular, line width=2pt, bottom color=yellow!30 [IntAct\\BioGRID, bottom color=green!20]]
           [Integrator, line width=2pt,  bottom color=yellow!30 [OmniPath, bottom color=green!20]]
          ]
         [Disease, bottom color=cyan!30
         [KEGG Disease\\DISEASES\\ DisGeNET\\OMIM\\GWAS Catalog, bottom color=green!20]
         ]
        [Compounds, bottom color=cyan!30
         [Chemicals, line width=2pt, bottom color=yellow!30 [ChEMBL\\PubChem, bottom color=green!20]]
         [Drugs, line width=2pt, bottom color=yellow!30 [DrugBank\\DrugCentral\\BindingDB\\RepoDB, bottom color=green!20]]
        ]
        [Pathway, bottom color=cyan!30
        [Reactome\\WikiPathways\\KEGG Pathways, bottom color=green!20]
        ]
        ]
      \end{forest}}
      \caption{Dataset Taxonomy}
      \label{fig:dataset-tax}
      \vskip -10pt
\end{figure*}

In this section we introduce some of the key resources providing information on crucial entities which should be included in a drug discovery KG: genes \& gene products, disease and drugs, as well as sources capturing the relationships between them via interactions, pathways and processes. A taxonomy of these datasets is presented in Figure \ref{fig:dataset-tax}. This list is not designed to be exhaustive, instead here we signpost some of the most popular and trusted ones and suggest what relations could be captured from them for a KG.

\emph{Note on tables:} Tables \ref{tab:kg-domain-specific}, \ref{tab:kg-dataset-protein}, \ref{tab:kg-dataset-interactions}, \ref{tab:kg-dataset-pathways}, \ref{tab:kg-dataset-diseases} and \ref{tab:kg-dataset-drugs} compare datasets on when they were first released, how regularly they are updated, ELIXIR core resource status and if free commerical use is allowed.

\subsection{Integrated Drug Discovery Resources}
\label{ssec:intergreated-datasets}

Table \ref{tab:kg-domain-specific} outlines resources which are tailored specifically for the drug discovery field. Typically these resources combine entity specific data sources and add additional information useful for the domain. These resources can also be useful as a reference point for some best practices with regards to data handling and integration.

\begin{table}[h!]
    \centering
    \renewcommand{\arraystretch}{1.2}
    \resizebox{\textwidth}{!}{
    \begin{tabular}{m{0.15\textwidth} C{0.11\textwidth} C{0.15\textwidth} C{0.12\textwidth} C{0.15\textwidth} C{0.16\textwidth} C{0.55\textwidth}} 
    \toprule
    \textbf{Dataset} & \textbf{First Released} & \textbf{Update Frequency} & \textbf{ELIXIR Core} & \textbf{Data Access} & \textbf{Commercial Use} & \textbf{Summary}\T\B \\ 
    \midrule \midrule

    Open Targets Platform & 2016 & $>$ Quarterly & \xmark & REST, Python, Flat files & \cmark & Resource focused for target discovery. Contains information from 20 data sources including UniProt, Reactome and ChEMBL. \\

    \midrule
    Pharos & 2014 & $\approx$ Monthly & \xmark & GraphQL & \cmark & Front end for the TCRD database for the drugable genome. Contains information from ChEMBL, STRING and UniProt. \\

    \bottomrule
    \end{tabular}}
    \caption{Overview of integrated drug discovery resources.}
    \label{tab:kg-domain-specific}
    \vskip -15pt
\end{table}

\textbf{Open Targets Platform.} The Open Targets Platform is a resource which collects various disparate data sources together, covering the key entities for target discovery including genes, drugs and diseases~\cite{koscielny2017open, carvalho2019open}. As of version 21.09, it contains data on 14K diseases and 27K targets. Each release contains detailed version information for the constituent datasets. Recently, Open Targets has been expanded with the addition of a Genetics portal~\cite{ghoussaini2020open}, for studying genetic variants and their relation to disease.

As Open Targets is specifically designed to link potential target genes/proteins to diseases, each link is provided with annotated associative evidence scores for a variety of evidence classes including genetic, drug and text mining. This information is aggregated into a final association score, indicating how associated a certain target-disease pair is overall. Thus far, Open Targets has not been integrated into any of the existing drug discovery KGs. However it is a prime resource, with clear scope to enrich a KG with pertinent target discovery information. Further, the various associative scores could be used to weight relations and provide some notion of confidence.

\textbf{Pharos.} The Pharos resource provides data integrations with a particular focus on the \emph{druggable genome}~\cite{nguyen2017pharos}. Pharos is the front-end access point, with the underlying data resource being the Target Central Resource Database (TCRD), part of a National Institutes of Health (NIH) program.

The information contained within Pharos could be used to provide links between proteins and diseases. However, it also contains detailed weighted protein-protein interactions. Additionally, Pharos contains various information types which could be used as entity features. For example, for a given protein entity, Pharos contains structural and expression information which could be transformed into a generic and task agnostic set of features.

\subsection{Gene and Gene Products}
\label{sec:proteinandgene}

Genes and gene products (i.e.,\ transcripts, proteins) are the key entities related to drug discovery and as such there are numerous rich public resources related to them. The gene and gene product datasets are summarised in Table \ref{tab:kg-dataset-protein}.

\begin{table}[ht!]
    \renewcommand{\arraystretch}{1.2}
    \centering
    \resizebox{\textwidth}{!}{
    \begin{tabular}{m{0.1\textwidth} C{0.11\textwidth} C{0.13\textwidth} C{0.12\textwidth} C{0.15\textwidth} C{0.16\textwidth} C{0.55\textwidth}} 
    \toprule
    \textbf{Dataset} & \textbf{First Released} & \textbf{Update Frequency}  & \textbf{ELIXIR Core} & \textbf{Data Access} & \textbf{Commercial Use} & \textbf{Summary}\T\B \\ 
    \midrule \midrule

    UniProtKB & 2003 & 8 Weeks & \cmark & REST, Python, Java, SPARQL & \cmark & Primary protein resource. Can be mined for protein-protein interactions and protein features. \\

    \midrule
    Ensembl & 1999 & 3 Months  & \cmark & REST, MySQL dump & \cmark & Primary source for gene and transcripts. Gene-gene and gene-disease relationships can be extracted, as well as many gene-based features. \\

    \midrule
    RNAcentral & 2014 & 3--6 Months & \xmark & REST, Flat file & \cmark & One of the primary  sources of non-coding transcript data.\\

    \midrule

    Entrez Gene & 2003 & Daily & \xmark & Flat file & \cmark & Another primary gene data resource. Used in existing KGs for gene entity annotations. \\

    \bottomrule
    \end{tabular}}
    \caption{Primary data sources relating to Genes and Proteins.}
    \label{tab:kg-dataset-protein}
    \vskip -15pt
\end{table}

\textbf{UniProt.} UniProt is a collection of protein sequence and functional information started in its current form in 2003 and provides three core databases: UniProtKB, UniParc, UniRef~\cite{apweiler2004uniprot}. The UniProtKB database is the main protein resource and comprises two different resources: Swiss-Prot and TrEMBL~\cite{apweiler2004uniprot}. Swiss-Prot contains the expert annotated and curated protein information, whilst TrEMBL stores the automatically extracted information. Thus, as of UniProt version 2020\_6, TrEMBL contains a greater volume of entities at 195M versus the 563K entities in Swiss-Prot.

\textbf{Ensembl.} Ensembl is primarily a data resource for genetics from the EBI, covering many different species~\cite{yates2020ensembl}. It provides detailed information on gene variants, transcripts and position in the overall genome. This data is extracted via an automated annotation process which considers only experimental evidence.

\textbf{RNAcentral.} RNAcentral is an integrative resource of non-coding RNA transcripts from 28 expert databases covering 296 organisms~\cite{sweeney2019}.
It provides the sequence information, Gene Ontology annotations, RNA family definitions, and identifier mappings between the many RNA transcript and gene controlled vocabularies that are necessary to semantically  harmonize the nomenclature used across RNA interaction databases such as miRTarBase~\cite{chou2016}.

\textbf{Entrez Gene.} Entrez Gene is the database of the NCBI which provides gene-specific information~\cite{maglott2005entrez}. Gene can be viewed as an integrated resource of gene information, incorporating information from numerous relevant resources. As such, it contains a mix of curated and automatically extracted information~\cite{brown2015gene}. Owing to its status as an integrator of relevant resources, it provides the GeneID system, a unique integer associated to each catalogued gene. The GeneID can be useful as a translation service between other resources and is used by many of the existing KGs (see Section \ref{sec:exisiting_kgs}) as the primary ID for its gene entities.

\subsubsection{Gene and Gene Products Resource Comparison} Table \ref{tab:kg-dataset-protein-compare} summarises the potential types of relations and features which could be extracted from the gene and gene product resources. The table highlights that many of these resources contain rich information which could be mined for gene level features, be that from the gene or protein sequence (the sequence of nucleic or amino acids represented as base pair letters which can be mined to form a representation~\cite{rives2021biological}), structure (the structure the protein forms once folded) or expression level (to what level is the gene expressed in different tissue types). The table also shows these resources to be good for extracting gene or protein interactions, as well as links to functional annotations via links to Gene Ontology (GO).

\begin{table}[h!]
    \centering
    \resizebox{0.85\textwidth}{!}{
    \begin{tabular}{lC{0.1\textwidth}C{0.1\textwidth}C{0.1\textwidth}G{0.15\textwidth} G{0.15\textwidth} G{0.15\textwidth}} 
    \toprule
    & \multicolumn{3}{c}{\textbf{Potential Relations}} & \multicolumn{3}{c}{\textbf{Potential Features}} \\
    \cline{2-7}
    \textbf{Dataset}  & \textbf{G/GP-G/GP} & \textbf{Gene-Protein} & \textbf{G/GP-GO} & \textbf{Sequence} & \textbf{Structure} & \textbf{Expression} \T\B \\
    \midrule \midrule

    UniProtKB    & \cmark & - & \cmark & \cmark & \cmark & \cmark \\
    Ensembl      & - & \cmark & - & \cmark & \cmark & \cmark \\
    RNAcentral   & - & - & \cmark & \cmark & \cmark & - \\
    Entrez Gene  & \cmark & - & - & \cmark & \cmark & \cmark\\

    \bottomrule
    \end{tabular}}
    \caption{Comparing gene (G) \& gene product (GP) resources on what relational information and entity-level features they provide.}
    \label{tab:kg-dataset-protein-compare}
    \vskip -15pt
\end{table}

\subsection{Interactions, Pathways and Biological Processes}

In this section, we detail the resources specialising in the linking of the entities through interaction, processes and pathways. The interaction resources are presented in Table \ref{tab:kg-dataset-interactions}, whilst the processes and pathways resources are detailed in Table \ref{tab:kg-dataset-pathways}

\begin{table}[ht!]
    \centering
    \renewcommand{\arraystretch}{1.2}
    \resizebox{\textwidth}{!}{
    \begin{tabular}{m{0.1\textwidth} C{0.11\textwidth} C{0.13\textwidth} C{0.12\textwidth} C{0.15\textwidth} C{0.16\textwidth} C{0.55\textwidth}} 
    \toprule
    \textbf{Dataset} & \textbf{First Released} & \textbf{Update Frequency}  & \textbf{ELIXIR Core} & \textbf{Data Access} & \textbf{Commercial Use} & \textbf{Summary}\T\B \\ 
    \midrule \midrule

    STRING & 2003 & Monthly  & \cmark & REST, Flat file, edgelist & \cmark & One of the most commonly used sources for physical and functional protein-protein interactions in existing KGs. \\

    \midrule

    BioGRID & 2003 & Monthly  & \xmark & REST, Flat file, edgelist, Cytoscape & \cmark& Contains interactions between gene, protein and chemical entities with could be included directly in a KG.\\

    \midrule

    IntAct & 2003 & Monthly & \cmark & Flat file & \cmark & Contains molecular reactions between gene, protein and chemical entities. Uses UniProt for identifiers. \\

    \midrule

    OmniPath & 2016 & $>$ Annually & \xmark & REST, Flat file, Cytoscape, Python, R & \cmark & An integrator of interaction resources that could be included in a KG via its RDF version. \\

    \bottomrule
    \end{tabular}}
    \caption{Primary data sources relating to interactions.}
    \label{tab:kg-dataset-interactions}
    \vskip -15pt
\end{table}

\subsubsection{Interaction Resources}
\label{ssec:interaction}
\hfill \break

\textbf{STRING.} The Search Tool for the Retrieval of Interacting Genes/Proteins (STRING) captures protein-protein interactions from over 5K different organisms~\cite{szklarczyk2019string}. The interactions in STRING are taken from a range of sources, including curated ones taken directly from experimental data and those which are mined from the literature using NLP techniques.

\textbf{BioGRID.} The Biological General Repository for Interaction Datasets (BioGRID) is a resource maintained by a range of international universities which specialises in collecting information regarding the interactions between biological entities including proteins, genes and chemicals~\cite{stark2006biogrid}.

\textbf{IntAct \& MINT.} IntAct is a database of molecular interactions maintained by the EBI~\cite{hermjakob2004intact}. Protein entities in IntAct are represented using UniProt IDs allowing for cross-resource linking. IntAct is closely linked with Molecular Interaction database (MINT), another core resource providing various interaction types between proteins~\cite{licata2012mint}.

\textbf{OmniPath.} OmniPath is a comparatively new resource for biological signalling pathway information with a focus on humans and rodents~\cite{turei2016omnipath}. It integrates over 100 literature curated data resources containing information on signalling pathways, this including many datasets covered in this present review such as IntAct, Reactome and ChEMBL.

\subsubsection{Interactions Resource Comparison} Table \ref{tab:kg-dataset-interaction-compare} summarises the potential relations types and features which could be extracted from the interaction resources. The table shows these data sources to be a rich potential for mining gene-gene or protein-protein interactions, with resources like IntAct and BioGrid also being suitable for extracting relations between gene products and compounds. All of the resources are also suitable for extracting features from the including weightings between the interactions or by providing different types or levels of interaction.

\begin{table}[h!]
    \centering
    \resizebox{0.85\textwidth}{!}{
    \begin{tabular}{lC{0.1\textwidth}C{0.1\textwidth}C{0.1\textwidth}C{0.1\textwidth}C{0.15\textwidth} G{0.15\textwidth} G{0.15\textwidth}} 
    \toprule
    & \multicolumn{5}{c}{\textbf{Potential Relations}} & \multicolumn{2}{c}{\textbf{Potential Features}} \\
    \cline{2-8}
    \textbf{Dataset}  & \textbf{Gene-Gene} & \textbf{Protein-Protein} & \textbf{Gene-Protein}  & \textbf{Gene-Drug} & \textbf{Protein-Drug} & \textbf{Types} & \textbf{Weightings} \T\B \\
    \midrule \midrule

    STRING  & - & \cmark & - & - & - & \cmark & \cmark \\
    BioGRID & \cmark & - & - & \cmark & - & \cmark & \cmark \\
    IntAct & - & \cmark & \cmark & - & \cmark  & \cmark & \cmark \\
    OmniPath & \cmark & \cmark  & \cmark & \cmark & \cmark & \cmark & \cmark \\
    \bottomrule
    \end{tabular}}
    \caption{Comparing interaction resources on what relational information and features they provide.}
    \label{tab:kg-dataset-interaction-compare}
    \vskip -15pt
\end{table}

\subsubsection{Pathway Resources}
\label{ssec:pathway}

Pathway resources comprise expert-curated subsets of interactions that are relevant for a given biological processes (e.g., apoptosis) or pathogenic mechanisms that lead to disease.
There are implicit biases in the definitions which have been shown to be mitigated by harmonizing and combining their definitions when possible~\cite{mubeen2019}.

\begin{table}[ht!]
    \renewcommand{\arraystretch}{1.2}
    \centering
    \resizebox{\textwidth}{!}{
    \begin{tabular}{m{0.1\textwidth} C{0.11\textwidth} C{0.13\textwidth}  C{0.12\textwidth} C{0.15\textwidth} C{0.16\textwidth} C{0.55\textwidth}} 
    \toprule
    \textbf{Dataset} & \textbf{First Released} & \textbf{Update Frequency}  & \textbf{ELIXIR Core} & \textbf{Data Access} & \textbf{Commercial Use} & \textbf{Summary}\T\B \\ 
    \midrule \midrule

    Reactome & 2003 & $>$ Annually  & \cmark & Neo4J, Flat files & \cmark & A core resource for pathways and reactions. Amiable for graph representation and already included in several KGs. \\

    \midrule

    WikiPathways & 2008 & Monthly & \xmark & REST, SPARQL, RDF, Python, R, Java & \cmark & A crowdsourced collection of pathway resources. Also provided in graph amiable formats. \\
    \midrule

    KEGG Pathways & 1995 & Bi-Annually & \xmark & REST, R, Python & \xmark & A highly influential resource for pathways. Free use is limited to academic work only. \\

    \bottomrule
    \end{tabular}}
    \caption{Primary data sources relating to pathways and processes.}
    \label{tab:kg-dataset-pathways}
    \vskip -15pt
\end{table}

\hfill \break

\textbf{Reactome.} Reactome is a large and detailed resource comprising biological reactions and pathways collected across multiple species including those from several model organisms and humans~\cite{jassal2020reactome}.  Resources from Reactome have also already been included in existing KG resources like Hetionet.

\textbf{WikiPathways.} The WikiPathways project explores the use of crowdsourcing for community curation of pathway and interaction resources~\cite{slenter2018wikipathways}. Owing to its crowdsourced nature, domain scientists can add new and edit existing information to ensure better overall quality and it has been designed to complement existing resources such as Reactome.

\textbf{KEGG Pathways.} The Kyoto Encyclopedia of Genes and Genomes (KEGG) provides a database of manually curated pathways covering metabolism, cellular processes, diseases, drug pathways, genetic information processing, environmental information processing, and organismal systems~\cite{kanehisa2017kegg}.
However the data licensing prevents redistribution or bulk access.
Despite this, it remains a highly influential bioinformatics resource.

\subsubsection{Pathway Resource Comparison} Table \ref{tab:kg-dataset-pathway-compare} summarises the potential types of relations and features which could be extracted from the pathway resources. The table shows all resources can be mined for gene-pathway links, with Reactome and KEGG Pathway also providing links from drugs to affected pathways. The table also highlights how all of the resources contain text descriptions of the pathways, as well as a graph-based representation which could further be mined for features.

\begin{table}[h!]
    \centering
    \resizebox{0.7\textwidth}{!}{
    \begin{tabular}{lC{0.1\textwidth}C{0.1\textwidth}C{0.1\textwidth} G{0.15\textwidth} G{0.15\textwidth}} 
    \toprule
    & \multicolumn{3}{c}{\textbf{Potential Relations}} & \multicolumn{2}{c}{\textbf{Potential Features}} \\
    \cline{2-6}
    \textbf{Dataset}  & \textbf{Protein-Protein} & \textbf{Gene-Pathway} & \textbf{Drug-Pathway} & \textbf{Graph Representation} & \textbf{Text Description} \T\B \\
    \midrule \midrule

    Reactome  & \cmark & \cmark & \cmark  & \cmark & \cmark \\
    WikiPathways  & - & \cmark & - & \cmark & \cmark \\
    KEGG Pathways  & - & \cmark & \cmark  & \cmark & \cmark \\
    \bottomrule
    \end{tabular}}
    \caption{Comparing pathway resources on what relational information and features they provide.}
    \label{tab:kg-dataset-pathway-compare}
    \vskip -15pt
\end{table}

\subsection{Diseases}
\label{ssec:diseases}

We now detail the resources whose primary focus is providing information about diseases. These resources are detailed in Table \ref{tab:kg-dataset-diseases}. 

\begin{table}[ht!]
    \renewcommand{\arraystretch}{1.2}
    \centering
    \resizebox{\textwidth}{!}{
    \begin{tabular}{m{0.1\textwidth} C{0.11\textwidth} C{0.13\textwidth}  C{0.12\textwidth} C{0.15\textwidth} C{0.16\textwidth} C{0.55\textwidth}} 
    \toprule
    \textbf{Dataset} & \textbf{First Released} & \textbf{Update Frequency} & \textbf{ELIXIR Core} & \textbf{Data Access} & \textbf{Commercial Use} & \textbf{Summary}\T\B \\ 
    \midrule \midrule

    KEGG DISEASE & 2008 & Monthly  & \xmark & REST, Flat file & \xmark  & A comprehensive disease resource for viewing disease as part of a biological system. Access is restricted for industrial use. \\

    \midrule

    DISEASES & 2015 & Daily  & \xmark & Flat file & \cmark  & A resource detailing links between genes and diseases. Already commonly used in drug discovery KGs.\T\B \\

    \midrule

    DisGeNET & 2010 & Annually & \xmark & REST, SPARQL, SQL, Flat tile, R, Cytoscape & \xmark  & One of the most frequently used disease sources in existing KGs. Contains a mix of resources including experimental and text-mined data.\\

    \midrule

    OMIM & 1987 & Daily  & \xmark & REST, Flat file & \xmark & One of the oldest disease databases, focusing upon mendelian disorders. Can provide gene-disease relationships. \\

    \midrule

    GWAS Catalog & 2008 & Biweekly & \xmark & REST, Flat file & \cmark & Contains the results from GWAS studies which could be used to provide less studied links between genes and diseases into a KG. \\

    \bottomrule
    \end{tabular}}
    \caption{Primary data sources relating to disease.}
    \label{tab:kg-dataset-diseases}
    \vskip -15pt
\end{table}

\textbf{KEGG DISEASE.} The KEGG DISEASE database is part of the larger KEGG resource, in which diseases are modelled as perturbed states of the molecular network system~\cite{kanehisa2007kegg}. Each disease entry contains information of the perturbants to this system including genetic and environmental factors of diseases, as well as drugs. Different types of diseases, including single-gene (monogenic) diseases, multifactorial diseases, and infectious diseases, are all treated in a unified manner by accumulating such perturbants and their interactions. Users must register to gain access and industry users are charged for use.

\textbf{DISEASES.} DISEASES is a dataset designed to integrate evidence on disease-gene associations from automatic text mining, manually curated literature, cancer mutation data, and genome-wide association studies~\cite{pletscher2015diseases}, with clear indication given as to whether the data is curated or exacted from text. 

\textbf{DisGeNET.} The DisGeNET resource integrates a variety of data sources from expert curated repositories, GWAS catalogues, animal models and the scientific literature~\cite{pinero2015disgenet}.. The data is integrated from four primary source types including expert curated databases and inferred associations.

\textbf{Online Mendelian Inheritance in Man.} Online Mendelian Inheritance in Man (OMIM) is a comprehensive, authoritative compendium of human genes and genetic phenotypes, with particular focus on the molecular relationship between genetic variation and phenotypic expression~\cite{hamosh2000online}. However access is controlled and users must register with OMIM in order to be able to download it.

\textbf{GWAS Catalog.} The Genome-Wide Association Studies (GWAS) performed in the literature provide an unprecedented opportunity to investigate the impact of common variants on complex diseases. The GWAS Catalog provides a consistent, searchable and freely available database of Single Nucleotide Polymorphism (SNP) to trait associations which are extracted from both published and unpublished GWA studies~\cite{buniello2019nhgri}. The data inside the GWAS Catalog is taken from studies found in the literature which are curated by experts before being added.

\subsubsection{Disease Resource Comparison} Table \ref{tab:kg-dataset-disease-compare} summarises the potential relations and features which could be extracted from the disease resources. The table shows that, unsurprisingly establishing gene to disease links is the primary focus of these resources. However, KEGG DISEASE could also be used to extract links from disease to both drugs and pathways, whilst DisGeNET also provides disease-disease similarity links. All of the resources provide some level of evidence for the links, whilst KEGG DISEASE and OMIM contain text descriptions which could be mined for features. 

\begin{table}[h!]
    \centering
    \resizebox{0.85\textwidth}{!}{
    \begin{tabular}{lC{0.1\textwidth}C{0.1\textwidth}C{0.1\textwidth}C{0.15\textwidth} G{0.15\textwidth} G{0.15\textwidth}} 
    \toprule
    & \multicolumn{4}{c}{\textbf{Potential Relations}} & \multicolumn{2}{c}{\textbf{Potential Features}} \\
    \cline{2-7}
    \textbf{Dataset}  & \textbf{Disease-Disease} & \textbf{Disease-Gene}  & \textbf{Disease-Drug} & \textbf{Disease-Pathway} & \textbf{Text Description} & \textbf{Evidence} \T\B \\
    \midrule \midrule

    KEGG DISEASE    & - & \cmark &  \cmark & \cmark & \cmark & \cmark \\
    DISEASES        & - & \cmark & - & - & - & \cmark \\
    DisGeNET        & \cmark & \cmark  & - & - & - & \cmark \\
    OMIM            & - & \cmark & - & - & \cmark & \cmark \\
    GWAS Catalog    & - & \cmark & - & - & - & \cmark \\
    \bottomrule
    \end{tabular}}
    \caption{Comparing disease resources on what relational information and entity-level features they provide.}
    \label{tab:kg-dataset-disease-compare}
    \vskip -15pt
\end{table}

\subsection{Drugs and Compounds}
\label{ssec:compounds}

We will now detail datasets containing information relating to drugs and compounds. This includes information on the relationships between the drugs and the targets or diseases as well other information such as potential adverse side effects or drug-drug interactions. These resources are detailed in Table \ref{tab:kg-dataset-drugs}.

\begin{table}[ht!]
    \renewcommand{\arraystretch}{1.2}
    \centering
    \resizebox{\textwidth}{!}{
    \begin{tabular}{m{0.1\textwidth} C{0.11\textwidth} C{0.13\textwidth} C{0.12\textwidth} C{0.15\textwidth} C{0.16\textwidth} C{0.55\textwidth}} 
    \toprule
    \textbf{Dataset} & \textbf{First Released} & \textbf{Update Frequency}  & \textbf{ELIXIR Core} & \textbf{Data Access} & \textbf{Commercial Use} & \textbf{Summary}\T\B \\ 
    \midrule \midrule

    ChEMBL & 2009 & $>$ Annually  & \cmark & REST, SQL dump, SPARQL & \cmark & One of the primary resources for drug-like molecules. Could provide relational information between gene and drugs. \\

    \midrule

    PubChem & 2004 & As Sources Are & \xmark & REST, Flat file, SPARQL & \cmark & A comprehensive integrator of other chemical resources provided in RDF format, enabling easy incorporation into a KG. \\

    \midrule

    DrugBank & 2006 & $>$ Annually & \xmark & REST, Flat file & \xmark & A rich source of drug, disease and gene information. Free use is limited to academic work only. \\

    \midrule

    DrugCentral & 2016 & Annually  & \xmark & SQL, Flat file & \cmark & A collection of drug information extracted from literature and other sources. A potential source of drug features. \\

    \midrule

    BindingDB & 1995 & Weekly & \xmark & REST, Flat file & \cmark & A data resource of target protein and compound information. Already incorporated in existing KGs.\ \\

    \midrule

    RepoDB & 2017 & No Set Schedule  & \xmark & Flat file & \cmark & A resources of drug to disease links containing both successful and failed examples. A rare source of negative information.\ \\

    \bottomrule
    \end{tabular}}
    \caption{Primary data sources relating to drugs.}
    \label{tab:kg-dataset-drugs}
    \vskip -15pt
\end{table}

\textbf{ChEMBL.} The ChEMBL dataset is one of the primary resources containing information on drug-like molecules and compounds~\cite{mendez2019chembl}. The records captured in the database are taken from the literature and curated before being added. ChEMBL has been included in many other integrated resources such as OpenTargets and Pharos.

\textbf{PubChem.} The PubChem is a resource collecting information on chemical molecules maintained by the NCBI~\cite{kim2016pubchem}. PubChem can be considered an integrator resource, aggregating over 700 disparate resources including UniProt, ChEMBL and Reactome.

\textbf{DrugBank.} The DrugBank database can be considered both a bioinformatics and a cheminformatics resource, thus containing information on drugs and potential targets~\cite{wishart2008drugbank}.

\textbf{DrugCentral.} DrugCentral is a resource containing information on drugs and other pharmaceuticals~\cite{ursu2016drugcentral}. The database focuses upon collecting information on FDA and EMA approved drugs, with the information being collected and curated from the literature and drug-labels, as well as from external sources.

\textbf{BindingDB.} BindingDB is a resource containing information primarily on the interactions between potential drug-target proteins and drug-like molecules~\cite{chen2001bindingdb}. BindingDB contains information taken from a range of sources including information curated from the literature, as well as from ChEMBL. BindingDB contains information which could be used to add relations between compound, protein and pathway entities, as well as potentially some compound-based feature information.

\textbf{RepoDB.} RepoDB is a smaller and more focused resource containing information more suitable for drug repositioning than many of the ones highlighted thus far. It focuses upon providing drug to disease links, however it not only provides information about approved drugs, but also on drugs which failed at various stages of clinical trails or that have been withdrawn~\cite{brown2017standard}. This is interesting when considering knowledge graphs, as the data could be used to provide negative edges between drugs or diseases.

\subsubsection{Drug Resource Comparison} Table \ref{tab:kg-dataset-drug-compare} summarises potential relations and features which could be extracted from the drug resources. The table shows all the resources focus on providing links between drugs and genes, with PubChem and DrugBank being sources of drug-drug interactions and DrugCentral providing potential drug-disease linkages. Almost all of the resources provide compand structure information (Usually in a string-based format called SMILES which can be used to learn a representation~\cite{hirohara2018convolutional}) and numerical attributes (molecular weight for example). KEGG DISEASE and OMIM also provides text based descriptions of the drugs which could be mined. 

\begin{table}[h!]
    \centering
    \resizebox{0.85\textwidth}{!}{
    \begin{tabular}{lC{0.1\textwidth}C{0.1\textwidth}C{0.1\textwidth}C{0.15\textwidth} G{0.15\textwidth} G{0.15\textwidth}G{0.15\textwidth}} 
    \toprule
    & \multicolumn{4}{c}{\textbf{Potential Relations}} & \multicolumn{3}{c}{\textbf{Potential Features}} \\
    \cline{2-8}
    \textbf{Dataset}  & \textbf{Drug-Drug} & \textbf{Drug-Gene}  & \textbf{Drug-Disease} & \textbf{Drug-Pathway} & \textbf{Text Description} & \textbf{Structure} & \textbf{Attributes} \T\B \\
    \midrule \midrule

    ChEMBL   & - & \cmark & - & - & - & \cmark & \cmark \\
    PubChem   & \cmark & \cmark & - & \cmark & \cmark & \cmark & \cmark \\
    DrugBank   & \cmark & \cmark & - & - & \cmark & \cmark & \cmark \\
    DrugCentral & - & \cmark & \cmark & - & - & \cmark & \cmark \\
    BindingDB &  - & \cmark & - & - & \cmark & \cmark & \cmark \\
    RepoDB &  - & \cmark & - & - & - & - & - \\
    \bottomrule
    \end{tabular}}
    \caption{Comparing drug resources on what relational information and entity-level features they provide.}
    \label{tab:kg-dataset-drug-compare}
    \vskip -15pt
\end{table}

\subsection{Dataset Evaluation}

This section summarises the key comparison points we have identified in our consideration of primary domain-specific datasets.

\begin{table}[ht!]
    \centering
    \resizebox{0.95\textwidth}{!}{
    \begin{tabular}{lC{0.1\textwidth}C{0.2\textwidth}C{0.15\textwidth} C{0.15\textwidth} G{0.15\textwidth} G{0.15\textwidth}G{0.18\textwidth}} 
    \toprule
    & \multicolumn{4}{c}{\textbf{Data Sources}} & \multicolumn{3}{c}{\textbf{Annotations}} \\
    \cline{2-5}
    \textbf{Dataset} & \textbf{Expert Curated} & \textbf{Experimental Evidence} & \textbf{Predicted \& Automated} & \textbf{Integrator Resource} & \textbf{Provenance} & \textbf{Confidence} & \textbf{Directionality} \T\B \\ 
    \midrule \midrule

    UniProtKB  & \cmark & \cmark & \cmark & - & \cmark & \cmark & Undirected \\
    Ensembl  & - & \cmark & \cmark & - & \cmark & - &  Undirected\\
    RNAcentral  & - & - & - & \cmark & \cmark & - & Mix  \\
    Entrez Gene  & \cmark  & \cmark & - & -& \cmark & - &  Undirected\\
    
    \midrule

    STRING  & \cmark & \cmark & \cmark & - & \cmark & \cmark & Undirected \\
    BioGRID & \cmark & \cmark & - & - & \cmark & \cmark & Mix \\
    IntAct &  \cmark & \cmark & - & - & \cmark & \cmark & Undirected\\
    OmniPath & - & - & - & \cmark & \cmark & - & Mix \\

    \midrule

    Reactome  & \cmark & \cmark & \cmark & - & \cmark & \cmark & Mix \\
    WikiPathways  & \cmark & \cmark & - & - & \cmark & - & Mix \\
    KEGG Pathways  & \cmark & \cmark & - & - & \cmark & - & Mix \\

    \midrule 

    KEGG DISEASE    & \cmark & \cmark & - & - & \cmark & - & Mix \\
    DISEASES        & \cmark & \cmark & \cmark & - & \cmark & \cmark & Undirected \\
    DisGeNET        & \cmark & \cmark & \cmark & - & \cmark & \cmark & Undirected \\
    OMIM            & \cmark & \cmark & - & - & \cmark & - & Undirected \\
    GWAS Catalog    & \cmark & \cmark & - & - & \cmark & \cmark & Undirected \\

    \midrule

    ChEMBL   & \cmark & \cmark & \cmark & - & \cmark & \cmark & Undirected \\
    PubChem   & \cmark & \cmark & - & \cmark & \cmark & - & Mix  \\
    DrugBank   & \cmark & \cmark & - & - & \cmark & - & Undirected \\
    DrugCentral & \cmark & \cmark & \cmark & - & \cmark & - & Undirected \\
    BindingDB & \cmark & \cmark & - & \cmark & \cmark & \cmark & Undirected \\
    RepoDB & - & \cmark & - & - & \cmark & - & Undirected  \\

    \bottomrule
    \end{tabular}}
    \caption{Comparing sources and annotations for the primary resources.}
    \label{tab:kg-data-trust}
    \vskip -20pt
\end{table}

\subsubsection{Data Trust} Table \ref{tab:kg-data-trust} highlights the different types of information in the resources, in addition to information pertaining to the the level of annotation available. Resources are compared as to whether they are curated by human experts, if information is taken from some form of experimental evidence or predicted and automated pipelines, and if the dataset contains information extracted from other primary resources. Resources are also compared if the province of the information is available (linking to the original manuscript or source), if any form of confidence weight is provided on the information and the directionality of potential edges that could be mined. The table shows that many of the covered resources have some level of human curation but it should be noted that this does not guarantee the accuracy of the information, as human bias and error can still be a factor. The table also highlights that predicted and automatically derived data is contained within many key resources such as STRING and DisGeNET, something to be cognisant of when including these in a KG. There are also various integrator resources available, like Omnipath and PubChem, which aggregate other primary datasets. Whilst caution is needed around potential replicated knowledge, they offer a way for KGs to incorporate diverse information from a single resource.

\subsubsection{Relation Mining} 
\begin{figure}[!t]
	\centering
    \resizebox{0.75\textwidth}{!}{
    {\tiny
    \begin{tikzpicture}

        \node[circle, draw, very thick, minimum size=1.1cm, fill=nyellow!70, align=center] at (-2, -0.5) (g) {Genes \& \\ Products};
        \node[circle, draw, very thick, minimum size=1.1cm, fill=ngreen!70] at (6.2, -0.5) (d) {Diseases};
        \node[circle, draw, very thick, minimum size=1.1cm, fill=nblue!90] at (2, -2) (c) {Drugs};
        \node[circle, draw, very thick, minimum size=1.1cm, fill=nred!50] at (2, 1.5) (p) {Pathways};

        \node[draw, rectangle, rounded corners, align=center, fill=nwhite] at (-3.7,-0.5) (gg) {\textbf{Gene-Gene}\\ UniProtKB, Ensembl, \\ RNACentral, Entrez, \\ STRING, BioGRID, \\ IntAct, OmniPath \\ Reactome};
        \node[draw, rectangle, rounded corners, align=center, fill=nwhite] at (-0.1,0.8) (gp) {\textbf{Gene-Pathway}\\ Reactome, WikiPathways, \\ KEGG Pathways};
        \node[draw, rectangle, rounded corners, align=center, fill=nwhite] at (2,3.3) (gd) {\textbf{Gene-Disease}\\ DISEASES, DisGeNET, OMIM, Gwas Catalog, KEGG DISEASE};
        \node[draw, rectangle, rounded corners, align=center, fill=nwhite] at (7.7, -0.5) (dd) {\textbf{Disease-Disease}\\ DisGeNET, \\ Disease Ontology, \\ MonDO};
        \node[draw, rectangle, rounded corners, align=center, fill=nwhite] at (4.2,0.6) (dp) {\textbf{Disease-Pathway}\\ KEGG DISEASE};
        \node[draw, rectangle, rounded corners, align=center, fill=nwhite] at (4.2,-1.4) (dc) {\textbf{Disease-Drug}\\ KEGG DISEASE, \\ DrugCentral};
        \node[draw, rectangle, rounded corners, align=center, fill=nwhite] at (-0.2,-1.4) (gc) {\textbf{Gene-Drug}\\ BioGRID, OmniPath, \\ ChEMBL, PubChem, \\ DrugBank, DrugCentral \\ BindingDB, RepoDB};
        \node[draw, rectangle, rounded corners, align=center, fill=nwhite] at (2,-3.2) (cc) {\textbf{Drug-Drug}\\ PubChem, \\ DrugBank};
        \node[draw, rectangle, rounded corners, align=center, fill=nwhite] at (2,-0.2) (cp) {\textbf{Drug-Pathway}\\ Reactome, \\ KEGG Pathways, \\ PubChem};
        \node[draw, rectangle, rounded corners, align=center, fill=nwhite] at (2,2.5) (pp) {\textbf{Pathway-Pathway}\\ Gene Ontology};

        \draw[-, thick, color=black, bend right=90] (g) to (gg); 
        \draw[-stealth, thick, color=black, bend right=90] (gg.south) to (g.south);

        \draw[-, thick, color=black, bend left=90] (d.north) to (dd.north); 
        \draw[-stealth, thick, color=black, bend left=90] (dd.south) to (d.south);

        \draw[-, thick, color=black, bend left=90] (c) to (cc); 
        \draw[-stealth, thick, color=black, bend left=90] (cc) to (c);

        \draw[-, thick, color=black, bend left=90] (p) to (pp); 
        \draw[-stealth, thick, color=black, bend left=90] (pp) to (p);

        \draw[-, thick, color=black] (g) to (gp); 
        \draw[-stealth, thick, color=black] (gp) to (p); 

        \draw[-, thick, color=black] (g) to (gc); 
        \draw[-stealth, thick, color=black] (gc) to (c); 

        \draw[-, thick, color=black] (d) to (dc); 
        \draw[-stealth, thick, color=black] (dc) to (c); 

        \draw[-, thick, color=black] (d) to (dp); 
        \draw[-stealth, thick, color=black] (dp) to (p); 

        \draw[-, thick, color=black] (c) to (cp); 
        \draw[-stealth, thick, color=black] (cp) to (p); 

        \draw[-, thick, color=black, bend left=30] (g) to (gd); 
        \draw[-stealth, thick, color=black, bend left=25] (gd) to (d); 
            
    \end{tikzpicture}
    }}
	\caption{Dataset usage for relations to link entity types in a simplified drug discovery knowledge graph schema.}
	\label{fig:kg-relent}
    \vskip -10pt
\end{figure}
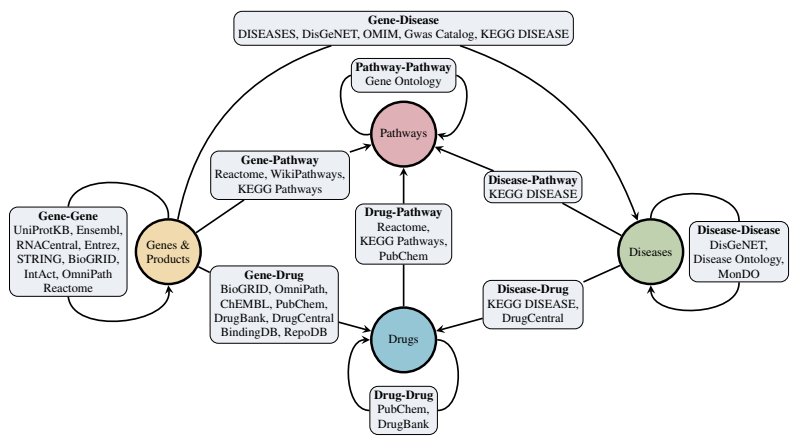

Figure \ref{fig:kg-relent} shows how the different resources covered in this review could be used to link key entities within a KG. The figure highlights how certain relation types are over-represented by the datasets, with Gene-Gene and Gene-Drug having many potential sources. Care should be taken to avoid duplicated edges if many of these resources are used in graph composition. The figure also highlights where information is lacking, with Disease-Pathway links only being present in one source. It also is interesting to note that many of the resources detailed here are already provided in some form that is amenable for ingestion into a knowledge graph -- either as edgelists or by providing RDF versions. This reduces the complexity of incorporating the resources as any issues arising from parsing and formatting process are avoided.

\subsubsection{Graph Enrichment}  There are many primary data sources which capture more information about key entities within drug discovery than just relational interactions. UniProtKB, for example, details numerous sequence and functional properties of proteins which may not be captured by relations alone. However, thus far, this wealth of information is under-explored and could be used to greatly enrich a KG with more domain knowledge. Of course this would come at the potential cost of some level of manual feature engineering being required -- an often complicated, domain specific and iterative process by itself, and one that much of the research into representation learning is attempting to avoid~\cite{bengio2013representation, mikolov2013distributed}. 

\subsubsection{Untapped Resources} Finally there are resources specific to drug discovery, such as OpenTargets and Pharos, which have thus far not been incorporated into any public KG. However, they are not currently provided in a format enabling easy incorporation into a KG, meaning that some manual conversion process is required. Yet they still hold great potential as a way to create a more drug discovery focused resource.

\section{Existing Biomedical Knowledge Graphs}
\label{sec:exisiting_kgs}

This section highlights the few existing knowledge graph datasets covering various aspects of the drug discovery process. These datasets often comprise graphs extracted from resources covering more primary information on the various relevant entities and relations. These datasets are interesting as they could form a good initial starting point for ML practitioners looking to test algorithms on suitable KGs. A selection of the most relevant resources is summarised in Table \ref{tab:kg-datasets}.

\begin{table}[ht!]
    \centering
    \resizebox{\textwidth}{!}{
    \begin{tabular}{p{0.2\textwidth} p{0.2\textwidth}  C{0.1\textwidth}  C{0.12\textwidth} C{0.1\textwidth} C{0.15\textwidth} C{0.15\textwidth}  C{0.15\textwidth} C{0.1\textwidth} C{0.1\textwidth}} 
    \toprule
    \textbf{KG Dataset}  & \textbf{Design Usecase} & \textbf{Entities} & \textbf{Triples} & \textbf{Entity Types} & \textbf{Relation Types} & \textbf{Contains Features} & \textbf{Constituent Datasets} & \textbf{Version Info} &  \textbf{Last Update}\T\B \\
    \midrule \midrule
    
    Hetionet~\cite{himmelstein2017systematic} & Repurposing & 47K  & 2.2M & 11 &  24 & \xmark & 29 & \xmark & 2017\T\B \\
    DRKG~\cite{drkg2020} & Repurposing & 97K  & 5.7M & 13 & 107 & molecular embeddings & 34 & \xmark & 2020\T\B \\
    BioKG~\cite{walsh2020biokg} & General & 105K &   2M & 10 &  17 & categorical & 13 & \xmark & 2020\T\B \\
    PharmKG~\cite{zheng2020pharmkg} & Repurposing/Target Prediction & 7.6K & 500K &  3 &  29 & continuous  &  7 & \xmark & 2020\T\B \\
    OpenBioLink~\cite{breit2020openbiolink} & Benchmark & 184K & 4.7M &  7 &  30 & \xmark      & 17 & \xmark & 2020\T\B \\
    Clinical Knowledge Graph~\cite{santos2020} & Personalised Medicine & 16M  & 220M & 35 &  57 & \xmark & 35 & \xmark & 2020 \T\B \\

    \bottomrule
    \end{tabular}}
    \caption{Pre-existing knowledge graphs suitable for use in various drug discovery applications.}
    \label{tab:kg-datasets}
    \vskip -20pt
\end{table}

\subsection{Biomedical Knowledge Graphs Overviews}

This section details graphs which we feel meet the criteria to be considered full knowledge graphs. 

\textbf{Hetionet v1.0.} One of the first attempts to create a holistic KG suitable for various tasks within drug discovery was Hetionet~\cite{himmelstein2017systematic}. Hetionet was developed as part of project Rephetio, a study looking at drug purposing through the use of KG-based approaches. The graph is publicly available\footnote{\url{https://het.io/}} and is provided as a Neo4j~\cite{have2013graph} dump, as well as in JSON and edge list. The underlying data is mined from sources including Entrez Gene~\cite{maglott2005entrez}, DrugBank~\cite{wishart2018drugbank}, DisGeNET~\cite{pinero2020disgenet}, Reactome~\cite{jassal2020reactome} and Gene Ontology~\cite{gene2008gene}. The thresholds for the edges are not included in the graph, instead the preselected values are detailed in the accompanying paper~\cite{himmelstein2017systematic}.

From the time of writing, Hetionet has not been updated since 2017, although a project called the Scalable Precision Medicine Oriented Knowledge Engine (SPOKE)~\cite{nelson2019integrating} looks to update Hetionet with extra data sources. However, to date, this updated resource has not been made publicly available, thus it has been excluded from our review.

\textbf{Drug Repurposing Knowledge Graph.} The Drug Repurposing Knowledge Graph (DRKG)~\cite{drkg2020} is a resource which builds upon Hetionet by integrating several additional data resources and was originally developed as part of a project for drug repurposing to target  COVID-19~\cite{ioannidis2020few}. The dataset is closely aligned with the Deep Graph Library (DGL) package for graph-based machine learning~\cite{wang2019dgl}, with pre-trained embeddings being provided from the package with the dataset. The data is publicly available\footnote{\url{https://github.com/gnn4dr/DRKG}.} and provided in edgelist format.

DRKG has enriched Hetionet with recent COVID-19 related data from STRING~\cite{szklarczyk2019string}, DrugBank~\cite{wishart2018drugbank} and GNBR~\cite{percha2018global}. DRKG also includes pre-computed GNN-based embeddings for molecules, however no other entities have associated features.

\textbf{BioKG.} BioKG is a project for integrating various biomedical resources and creating a KG from them~\cite{walsh2020biokg}. As part of the project, various tools are provided to enable a simplified KG construction process. A public pre-made version of the graph is available\footnote{\url{https://github.com/dsi-bdi/biokg}}, as well as the code for building it.

The data which makes up BioKG is taken from 13 different data sources, including UniProt~\cite{apweiler2004uniprot}, Reactome~\cite{jassal2020reactome}, OMIM~\cite{hamosh2000online} and Gene Ontology~\cite{gene2008gene}. One interesting aspect of BioKG is that a small number of categorical features are provided with some of the entities. For example, drug entities are enriched with information pertaining to any associated negative side effects.

\textbf{PharmKG.} The PharmKG project had the goal of designing a high quality general purpose KG and associated GNN-based model for use within the drug discovery domain~\cite{zheng2020pharmkg}. Table \ref{tab:kg-datasets} shows that compared to others highlighted in this section, PharmKG is compact, containing entities of just three types: chemical, gene and disease. 

The data is initially integrated from 7 sources including OMIM~\cite{hamosh2000online}, DrugBank~\cite{wishart2008drugbank}, PharmGKB~\cite{whirl2012pharmacogenomics}, Therapeutic Target Database (TTD)~\cite{chen2002ttd}, SIDER~\cite{kuhn2016sider}, HumanNet~\cite{hwang2019humannet} and GNBR~\cite{percha2018global}. A filtering process is then applied to ensure that only high quality knowledge is kept, for example by only including well studied genes. One unique aspect is that numerical features are provided with all the entities. Such features include chemical connectivity and other physiochemical features for the chemical entities, the use of BioBERT~\cite{lee2020biobert} to create features for the disease entities and a reduced expression matrix to create a feature vector for gene entities. The unfiltered PharmKG graph, as well as model code, is available to download\footnote{\url{https://github.com/MindRank-Biotech/PharmKG}}, however at the time of writing, neither the filtered graph or the entity features vectors have been released.

\textbf{OpenBioLink.} OpenBioLink (OBL) is a project to allow for easier and fairer comparison of KG completion approaches for the biomedical domain~\cite{breit2020openbiolink}. As part of the project, a benchmark KG has been created covering aspects of the drug discovery landscape. The dataset is publicly available\footnote{\url{https://zenodo.org/record/3834052}} and is provided in edgelist and RDF formats.

Data is taken from 17 datasets including STRING~\cite{szklarczyk2019string}, DisGeNET~\cite{pinero2020disgenet}, Gene Ontology~\cite{gene2008gene}, CTD~\cite{davis2019comparative}, Human Phenotype Ontology~\cite{kohler2019expansion}, SIDER~\cite{kuhn2016sider} and KEGG~\cite{kanehisa2010kegg}, among other resources. Of interest is that OpenBioLink contains additional \emph{true negatives} for a selection of relation types, meaning that this relation was explicitly detailed not to exist. This can be used to avoid the issues inherent with the choice of negative sampling strategy when training KG embedding models~\cite{zhang2019nscaching}.

\textbf{Clinical Knowledge Graph.} The Clinical Knowledge Graph (CKG) builds upon previous benchmark KGs but with additional focus on \textit{-omics} data.
Its relations come from 25 databases and 10 ontologies, many of which overlap with previous examples but notably include protein state information such as post-translational modifications from PhosphoSite~\cite{hornbeck2015}.
The CKG GitHub repository\footnote{\url{https://github.com/MannLabs/CKG/}} not only provides code for rebuilding the graph, but also tools for uploading it into Neo4J as well as visualization and exploration in Jupyter Notebooks.
However, CKG cannot redistribute many of its constituent datasets because of licensing restrictions, so the distributed version of the CKG is much smaller than stated in the manuscript.

\subsection{Comparative Analysis of KG Resources}
\label{ssec:comparison-kg}

We now present a comparative analysis of the KGs by considering graph composition choices, dataset usage and documentation levels.\footnote{Note that we exclude CKG from much of this analysis due to licensing limitations.} This analysis is undertaken to better understand the types of drug discovery problems each graph is suitable for addressing, as well as allowing interpretation of the level of trust that can be placed in each graph through exploration of dataset province. We believe this is the first time these resources have been compared and contrasted in the literature.

\begin{table}[ht!]
    \centering
    \resizebox{\textwidth}{!}{
    \begin{tabular}{lgggccggcgcgcc} 
    \toprule
    & \multicolumn{3}{c}{\textbf{Gene Products}} & \multicolumn{2}{c}{\textbf{Compounds}} & \multicolumn{2}{c}{\textbf{Disease}}  &  &  \multicolumn{1}{c}{} & \multicolumn{1}{c}{} &  \multicolumn{1}{c}{} \\
    \cline{2-8}
    \textbf{KG Dataset}  & \textbf{Gene} & \textbf{Proteins} & \textbf{Transcripts} & \textbf{Drugs} & \textbf{Chemicals} & \textbf{Disease} & \textbf{Genetic Disorder} & \textbf{Anatomy} & \textbf{Pathways} & \textbf{Side Effect} & \textbf{Symptoms}
    \T\B \\
    \midrule \midrule
    Hetionet                 & \cmark &    -    &    -    & \cmark &   -     & \cmark &    -    & \cmark & \cmark & \cmark & \cmark \\
    DRKG                     & \cmark &    -    &    -    & \cmark &   -     & \cmark &   -     & \cmark & \cmark & \cmark & \cmark\\
    BioKG                    &    -    & \cmark &    -    & \cmark &   -     & \cmark & \cmark &     -   & \cmark &     -   &- \\
    PharmKG                  & \cmark &    -    &    -    &        & \cmark & \cmark &    -    &    -    &        &   -     & -\\
    OpenBioLink              & \cmark &     -   &     -   & \cmark &  -      & \cmark &     -   & \cmark & \cmark &  -      &- \\
    CKG                      & \cmark & \cmark & \cmark & \cmark &    -    & \cmark &   -     &     -   & \cmark &   -     &- \B \\

    \bottomrule
    \end{tabular}}
    \caption{Comparison of a subset of entity types named across the knowledge graphs.}
    \label{tab:kg-entity}
    \vskip -20pt

\end{table}

\begin{figure}[!ht]
	\centering
	\resizebox{\textwidth}{!}{
	\begin{subfigure}[b]{0.4999999\textwidth}
		\centering

        \includegraphics[width=\textwidth]{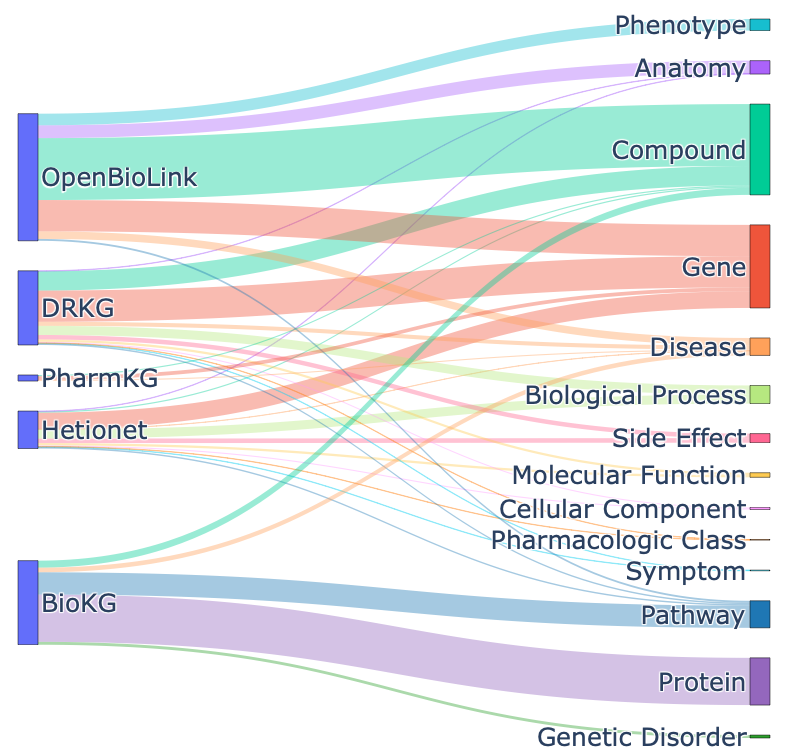}

	\caption{Entities.}\label{fig:kg-sankey-entity}
	\end{subfigure}
	\hfill
	\begin{subfigure}[b]{0.499999\textwidth}
		\centering		
        \includegraphics[width=\textwidth]{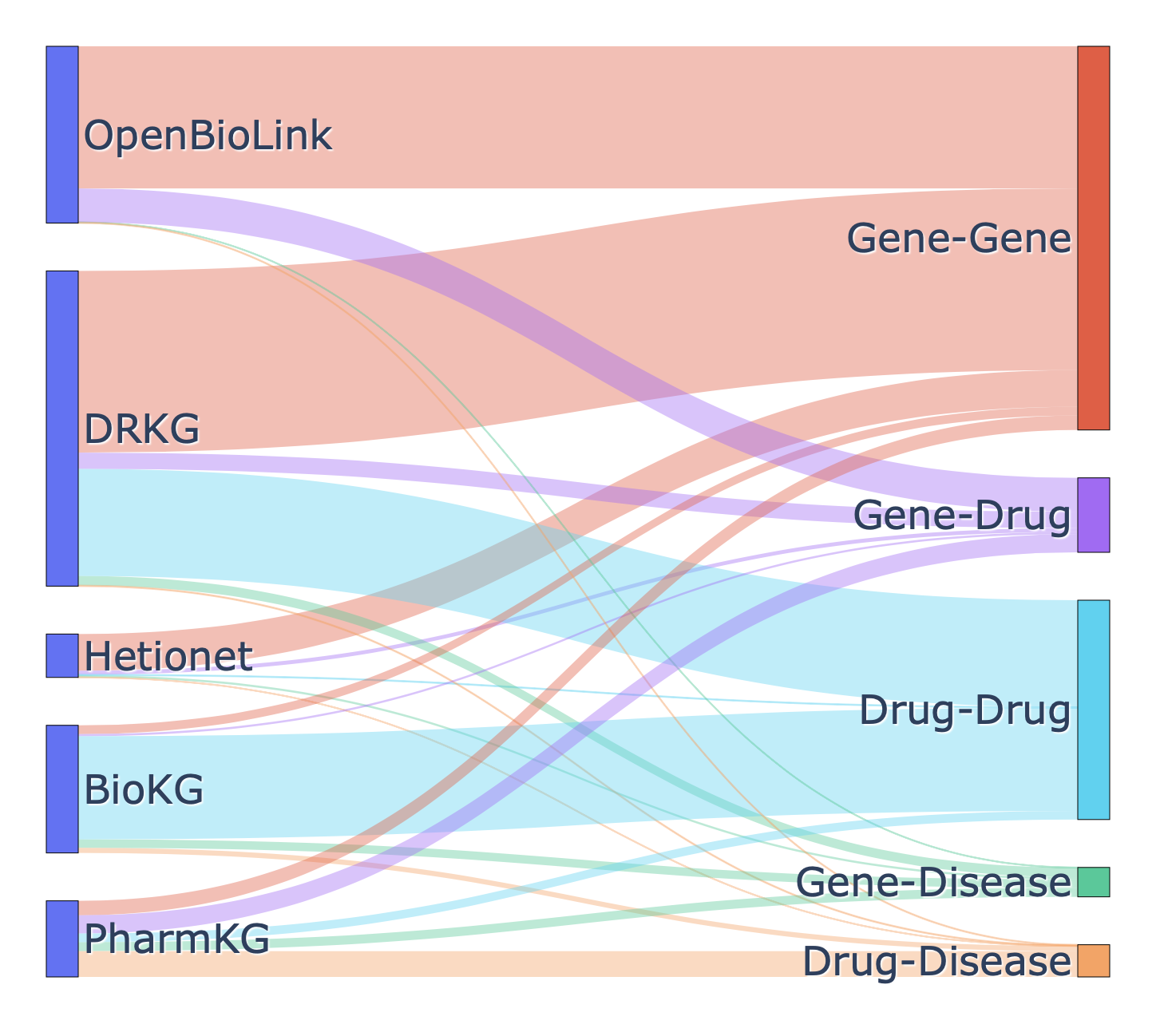}
	\caption{Relations.}\label{fig:kg-sankey-relation}
	\end{subfigure}}

	\caption{Sankey diagrams showing relationship between entity and relations in the KGs. Line thickness equates to entity volume. Note that for the relations, the value is the sum of all relation types between the two entities.}
	\vskip -10pt
	\label{fig:kg-sankey}
\end{figure}

\begin{table}[ht!]
    \centering
    \resizebox{0.6\textwidth}{!}{
    \begin{tabular}{lcccc} 
    \toprule
    \textbf{KG Dataset}  & \textbf{Gene/Products} & \textbf{Compound} & \textbf{Disease} & \textbf{Pathways} \T\B \\
    \midrule \midrule

    Hetionet    & Entrez GeneID & DrugBank AN & DOID & Custom\\
    DRKG        & Entrez GeneID & DrugBank AN & MeSH & Custom \\
    BioKG       & UniProt       & DrugBank AN & MeSH & Reactome/KEGG \\
    PharmKG     & Entrez GeneID & PubChem ID  & MeSH & \xmark \\
    OpenBioLink & Entrez GeneID & PubChem ID  & DOID & Reactome/KEGG \B \\

    \bottomrule
    \end{tabular}}
    \caption{Entity identifiers used in the different knowledge graphs. Multiple IDs being present means all are used as entity identifiers within the graph.}
    \vskip -15pt
    \label{tab:kg-entity-mappings}

\end{table}

\subsubsection{Graph Composition: Entities} Table \ref{tab:kg-entity} highlights which entity types are included in the KGs as well as offering a fine grained view of how larger concepts like gene-products are modelled, whilst Figure \ref{fig:kg-sankey-entity} shows the amount of these different entities present across the KGs. These show that the KGs take differing approaches to how entities are modelled and the volumes included, which in turn could determine for which task they are best suited. Overall the different KGs share only three common entities: gene products (be that genes or proteins), compounds and disease. As these are the core entities involved in drug discovery, this is no surprise. Pathways are also frequently included, with only PharmKG leaving them absent. BioKG and CKG are the only resources to model at the level of proteins instead of genes, whilst BioKG is the only resource to split genetic disorders from diseases. It can also be seen that Hetionet (and DRKG by virtue of it being an expanded Hetionet) captures more ancillary information compared to other KGs in the form of entities such as drug side effects, disease symptoms and various gene-level annotations, and thus might be well suited for tasks which could benefit from more fine-grained information. 

Table \ref{tab:kg-entity-mappings} highlights the identifiers used by the KGs for key entities. Typically entities of a certain type are represented using one of two choices of identifier, with pathways having the lowest level of consistency. Knowing which identifiers are used allows additional sources of information to be joined onto the graphs with greater ease.

\subsubsection{Graph Composition - Relations} Table \ref{tab:kg-relation-types} shows the number of different relationship types in the KGs between key entity pairs. The table highlights the different nuance with which the relationships are modelled. It is clear there is a large variation in what an edge between entity pairs is actually representing. However, note that the values for DRKG are inflated as the data source is captured in the relationship name. Considering the relation granularity can further help guide on KG task suitability. For example Hetionet and OpenBioLink both have multiple relationship types between gene entities, whilst other KGs have only one, perhaps indicating these to be good choices if a complex understanding of gene interaction is required. Whereas, OpenBioLink would not be the graph to use if interaction between drugs was crucial to the task as it has no drug-drug edges. Overall, some general trends are observable regarding relation modelling choices. For example, BioKG tends to use only a single relation type, whilst PharmKG, despite its smaller overall size, often choses to have multiple types. Additionally, it can be seen in Table \ref{tab:kg-relation-types} that drug entity pairs are consistently modelled as only a single relation type across the graphs.

Figure \ref{fig:kg-sankey-relation} displays the volume of each relation category contained within the graphs and shows there to be a marked difference between the KGs. For example, DRKG has a large number of both Gene-Gene and Drug-Drug edges in comparison to other types, whilst OpenBioLink choses to include more gene interactions and BioKG has a large number of drug interactions. Overall Gene-Gene and/or Drug-Drug relations form the majority in many of the KGs. This has the potential to cause issues for tasks like target discovery, which relies on gene to disease connections, as there tend to be fewer examples in the graph. Thus, any model training on top of these graphs will have fewer to learn from, potentially leading to suboptimal predictive performance.

\begin{table}[h!]
    \centering
    \resizebox{0.75\textwidth}{!}{
    \begin{tabular}{lccccc} 
    \toprule
    \textbf{KG Dataset}  & \textbf{Gene-Gene} & \textbf{Gene-Disease} & \textbf{Gene-Drug} & \textbf{Drug-Drug} & \textbf{Drug-Disease} \T\B \\
    \midrule \midrule

    Hetionet    & 3 & 3 & 3 & 1 & 2\\
    DRKG        & 32 & 15 & 34 & 2 & 10\\
    BioKG       & 1 & 1 & 5 & 1 & 1\\
    PharmKG     & 1 & 6 & 7 & 1 & 6 \\
    OpenBioLink & 10 & 1 & 10 & \xmark & 1\B \\

    \bottomrule
    \end{tabular}}
    \caption{The number of relation types between entities across the KGs.}
    \label{tab:kg-relation-types}
    \vskip -15pt
\end{table}

\subsubsection{Underlying Dataset Use} 
\begin{figure}[!h]
	\centering
    {
    \tiny
    \resizebox{0.7\textwidth}{!}{
    \begin{tikzpicture}

        \Vertex[IdAsLabel, shape=rectangle, size=1., x=2, RGB ,color={163,190,140}]{Hetionet}
        \Vertex[IdAsLabel, shape=rectangle, size=1., RGB ,color={163,190,140}]{OBL}
        \Vertex[IdAsLabel, shape=rectangle, size=1., x=-2, RGB ,color={163,190,140}]{DRKG}
        \Vertex[IdAsLabel, shape=rectangle, size=1., x=-4, RGB ,color={163,190,140}]{BioKG}

        \Vertex[IdAsLabel, size=1., x=2.4, y=3.5]{HID}
        \Vertex[IdAsLabel, size=1., x=3.5, y=3]{LINCS}
        \Vertex[IdAsLabel, size=1.3, x=4.5, y=1.5]{DISEASES}
        \Vertex[IdAsLabel, size=1.1, x=4.5, y=-1]{GWAS C}
        \Vertex[IdAsLabel, size=1.2, x=2.8, y=-3.5]{LabeledIn}
        \Vertex[IdAsLabel, size=1., x=3.5, y=-2]{MEDI}

        \Vertex[IdAsLabel, size=1., x=-3, y=-3]{GNBR}
        \Vertex[IdAsLabel, size=1., x=-1, y=2]{DGIdb}

        \Vertex[IdAsLabel, size=1., x=-6, y=2.5]{Uniprot}
        \Vertex[IdAsLabel, size=1., x=-7, y=0.5]{KEGG}
        \Vertex[IdAsLabel, size=1., x=-6.5, y=-1.5]{OMIM}

        \Vertex[IdAsLabel, size=1., x=-.2, y=-3.]{STITCH}

        \Vertex[IdAsLabel, size=1., x=-5, y=-3]{IntAct}
        \Vertex[IdAsLabel, size=1.2, x=1, y=3]{DisGeNET}
        \Vertex[IdAsLabel, size=1., x=-1.5, y=-3.5]{STRING}
        \Vertex[IdAsLabel, size=1.35, x=1, y=-3.5]{DrugCentral}
        \Vertex[IdAsLabel, size=1.2, x=-2, y=4]{DrugBank}


        \Edge[Direct, Math, label=\sigma, lw=1.5pt, RGB, color={235, 203, 139}](HID)(Hetionet)
        \Edge[Direct, Math, label=\sigma, bend=10, lw=1.5pt, RGB, color={235, 203, 139}](LINCS)(Hetionet)
        \Edge[Direct, Math, label=\sigma, lw=1.5pt, RGB, color={235, 203, 139}](STRING)(DRKG)
        \Edge[Direct, Math, label=\sigma, bend=-20, lw=1.5pt, RGB, color={235, 203, 139}](GNBR)(DRKG)
        \Edge[Direct, Math, label=\sigma, bend=-10, lw=1.5pt, RGB, color={235, 203, 139}](IntAct)(DRKG)
        \Edge[Direct, Math, label=\sigma, bend=-10, lw=1.5pt, RGB, color={235, 203, 139}](IntAct)(BioKG)
        \Edge[Direct, Math, label=\sigma, lw=1.5pt, RGB, color={235, 203, 139}](Uniprot)(BioKG)
        \Edge[Direct, Math, label=\sigma, lw=1.5pt, RGB, color={235, 203, 139}](STRING)(OBL)

        \Edge[Direct, Math, label=\mu, lw=1.5pt, RGB, color={191, 97, 106}](DISEASES)(Hetionet)
        \Edge[Direct, Math, label=\mu, lw=1.5pt, RGB, color={191, 97, 106}](GWAS C)(Hetionet)
        \Edge[Direct, Math, label=\mu, lw=1.5pt, RGB, color={191, 97, 106}](DisGeNET)(Hetionet)
        \Edge[Direct, Math, label=\mu, bend=-10, lw=1.5pt, RGB, color={191, 97, 106}](GNBR)(DRKG)
        \Edge[Direct, Math, label=\mu, bend=-10, lw=1.5pt, RGB, color={191, 97, 106}](KEGG)(BioKG)
        \Edge[Direct, Math, label=\mu, lw=1.5pt, RGB, color={191, 97, 106}](OMIM)(BioKG)
        \Edge[Direct, Math, label=\mu, bend=-11, lw=1.5pt, RGB, color={191, 97, 106}](DisGeNET)(OBL)

        \Edge[Direct, Math, label=\lambda, lw=1.5pt, RGB, color={163, 190, 140}](DrugCentral)(Hetionet)
        \Edge[Direct, Math, label=\lambda, bend=-10, lw=1.5pt, RGB, color={163, 190, 140}](LINCS)(Hetionet)
        \Edge[Direct, Math, label=\lambda, lw=1.5pt, RGB, color={163, 190, 140}](DGIdb)(DRKG)
        \Edge[Direct, Math, label=\lambda, bend=10, lw=1.5pt, RGB, color={163, 190, 140}](GNBR)(DRKG)
        \Edge[Direct, Math, label=\lambda, bend=10, lw=1.5pt, RGB, color={163, 190, 140}](IntAct)(DRKG)
        \Edge[Direct, Math, label=\lambda, bend=10, lw=1.5pt, RGB, color={163, 190, 140}](KEGG)(BioKG)
        \Edge[Direct, Math, label=\lambda, bend=10, lw=1.5pt, RGB, color={163, 190, 140}](IntAct)(BioKG)
        \Edge[Direct, Math, label=\lambda, lw=1.5pt, RGB, color={163, 190, 140}](STITCH)(OBL)

        \Edge[Direct, Math, label=\theta, lw=1.5pt, RGB, color={180, 142, 173}](DrugBank)(Hetionet)
        \Edge[Direct, Math, label=\theta, bend=10, lw=1.5pt, RGB, color={180, 142, 173}](DrugBank)(DRKG)
        \Edge[Direct, Math, label=\theta, bend=10, lw=1.5pt, RGB, color={180, 142, 173}](DrugBank)(BioKG)

        \Edge[Direct, Math, label=\gamma, lw=1.5pt, RGB, color={94, 129, 172}](LabeledIn)(Hetionet)
        \Edge[Direct, Math, label=\gamma, lw=1.5pt, RGB, color={94, 129, 172}](MEDI)(Hetionet)
        \Edge[Direct, Math, label=\gamma, lw=1.5pt, RGB, color={94, 129, 172}](GNBR)(DRKG)
        \Edge[Direct, Math, label=\gamma, bend=-10, lw=1.5pt, RGB, color={94, 129, 172}](DrugBank)(DRKG)
        \Edge[Direct, Math, label=\gamma, bend=-10, lw=1.5pt, RGB, color={94, 129, 172}](DrugBank)(BioKG)
        \Edge[Direct, Math, label=\gamma, lw=1.5pt, RGB, color={94, 129, 172}](DrugCentral)(OBL)

    \end{tikzpicture}}}
	\caption{The relationship between drug discovery knowledge graphs and underlying data sources. Relationships are presented for five major relation categories: Gene-Gene (\(\sigma\)), Gene-Disease (\(\mu\)), Gene-Drug (\(\lambda\)), Drug-Drug (\(\theta\)) and Drug-Disease (\(\gamma\)).}
	\label{fig:kg-dataset-relations}
    \vskip -15pt
\end{figure}
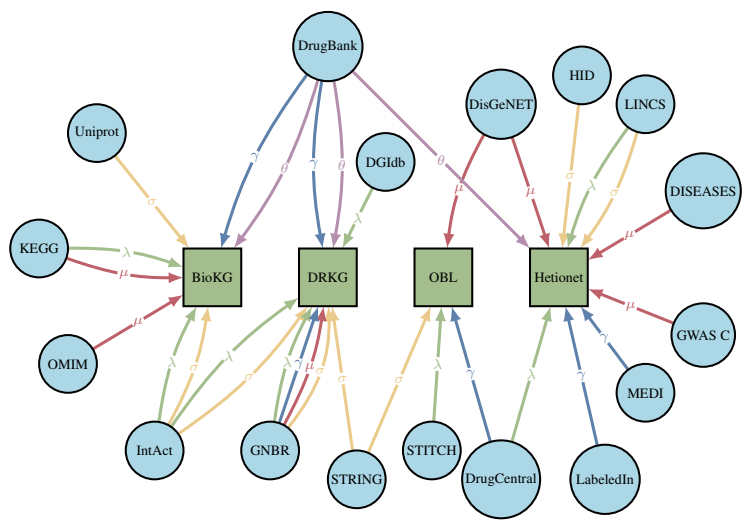

Figure \ref{fig:kg-dataset-relations} represents dataset use in the KGs for a series of key relationship types, where the typed edges indicate relations are taken from that dataset. PharmKG and CKG are missing as it was not possible to determine the data sources used for the relations. The figure shows that the KGs utilise many of the same underlying datasets, with DrugBank for example being used in the majority of the graphs, with often multiple relationship types being extracted from it.

Considering the difference in dataset usage, we can see some of the choices made during the composition pertaining to the KGs intended use cases. For example DRKG, extracts four different relationship types from the text mining-based GNBR~\cite{percha2018global} dataset, whilst no other KG uses any. The creators of DRKG must have deemed these lower-confidence edges useful for discovering potential repurposing candidates for COVID-19. It can also be seen that Hetionet tends to use multiple smaller datasets to build a single relation type, the gene-disease edges use three for example. Hetionet also differs in that its drug-disease edges don't come from larger aggregator resources DrugBank or DrugCentral used by the other KGs. In contrast, OpenBioLink has a one-to-one mapping between dataset and relation type and makes use of larger resources like STRING and STITCH, perhaps showing its intended benchmark use. 

Figure \ref{fig:kg-dataset-relations} also highlights some of the pitfalls of mining multiple resources for relations of the same type. DRKG extracts gene-gene interaction edges from both IntAct and STRING, which could result in duplicated edges being present as both datasets contain many of the same interactions. Without care during the composition and evaluation process, this could lead to situations where training edges used in a model could also potentially be used for evaluation. 
 
\subsubsection{Evaluation of Documentation Quality and Reproducibility} Table \ref{tab:kg-documentation} presents our evaluation of the documentation quality (be that from the original paper, supplementary material, code repository or website) and overall reproducibility of the KGs. The graphs are evaluated using the following criteria, where the documentation quality categories are scored from one to three - \emph{Schema Overview:} Is the graph schema design well explained and justified? A score of three means all entity and relation types detailed in full, two means the schema is outlined but not fully justified and one means only basic details are provided. \emph{Dataset Filtering:} Is there a clear description of how the underlying datasets were filtered? A score of three indicates filtering thresholds detailed enough for reproducibility, two means that some description is provided but not enough to reproduce the work and one that only a limited amount of information is provided. \emph{Relation Explanation:} Is the meaning behind relations well explained? A score of three indicates each relation type is fully explained and linked to the source dataset, two that either a full relation explanation or source dataset mapping was missing and one that no mappings were provided. \emph{Updates:} Are any future planned updates detailed? \emph{Data-Relation Mappings:} Is it possible to map edges directly back to the underlying data sources? \emph{Construction Code:} Is code available to construct the graph? \emph{Licence Info:} Are underling dataset licences detailed?

Table \ref{tab:kg-documentation} highlights that, despite being the oldest resource, Hetionet remains the KG with the highest overall level of documentation quality. Regarding reproducibility, two of the KGs did not provide code to recreate the graphs from the source datasets. It was also interesting to note that none of the resources provided any details on whether they would be updated going forward. Overall its clear that further work needs to be undertaken to improve documentation and reproducibility which will aid in both increasing trust and also ease of use of future KG resources.

\begin{table}[h!]
    \centering
    \resizebox{\textwidth}{!}{
    \begin{tabular}{lC{0.12\textwidth}C{0.12\textwidth}C{0.15\textwidth}cC{0.45\textwidth}G{0.2\textwidth} G{0.15\textwidth} G{0.1\textwidth}} 
    \toprule
    & \multicolumn{5}{c}{\textbf{Documentation}} & \multicolumn{3}{c}{\textbf{Reproducibility}} \\
    \cline{2-9}
    \textbf{KG Dataset}  & \textbf{Schema Overview} & \textbf{Dataset Filtering} & \textbf{Relation Explanation} & \textbf{Updates} & \textbf{Overvierw} & \textbf{Data-Relation Mappings} & \textbf{Construction Code} & \textbf{Licence Info} \T\B \\
    \midrule \midrule

    Hetionet    & \cmark\cmark\cmark & \cmark\cmark\cmark & \cmark\cmark\cmark & \xmark & Graph creation process well documented and design choices well explained. & \cmark & \cmark & \cmark \\

    DRKG        & \cmark\cmark & \cmark & \cmark\cmark & \xmark & Dataset usage well documented but graph creation details lacking. & \cmark & \xmark & \cmark \\

    BioKG       & \cmark\cmark & \cmark & \cmark\cmark & \xmark & Overall good, could be improved by more details on construction choices  & \cmark & \cmark & \cmark\\
    PharmKG     & \cmark & \cmark\cmark & \cmark\cmark & \xmark & Dataset to relation mapping not detailed but graph creation process explained. & \xmark & \xmark & \xmark\\
    OpenBioLink & \cmark\cmark\cmark & \cmark\cmark & \cmark\cmark & \xmark & Well documented but relations could be better explained. & \cmark & \cmark & \cmark\B \\

    \bottomrule
    \end{tabular}}
    \caption{Comparing documentation levels and reproducibility across the KGs. Documentation quality is scored on scale of 1 to 3.}
    \label{tab:kg-documentation}
    \vskip -20pt
\end{table}

\subsection{Shortcomings of Existing KG Resources}
\label{ssec:analysis-kg}

When looking at these existing KG as a whole, we can identify the following shortcomings:

\begin{itemize}

    \item \emph{Lack of Updates -} None of the detailed KGs have any form of maintenance or update schedule in place. This means they will become increasingly out of date as the underlying data resources continue to evolve.

    \item \emph{Lack of Detailed Documentation -} Some of the resources are not properly documented, missing clear justifications for some of the design choices, not including crucial information for reproducibility such as threshold information and lacking clear mappings back to the source datasets. This makes the graphs more challenging to use in tackling real-world problems.
   
    \item \emph{Lack of Features -} Almost all of these graphs do not provide any additional features for the entities or relations. Features that are provided are usually limited to only a small number of entities.
    
    \item \emph{No Dataset Version Information -} Many of the resources do not detail from which version or year of a certain dataset the information has been collected. 
    
\end{itemize}

\subsection{Other Associated Graphs \& Projects}

\subsubsection{Other Graph Datasets} Despite not being full KGs, there are a number of other graphs of note in the domain. For example, the Stanford Biomedical Network Dataset Collection (BioSNAP)\footnote{\url{https://snap.stanford.edu/biodata/}} is collection of homogeneous graphs, where each graph captures the interactions between two (from a total of ten) entity types such as gene-gene or gene-disease~\cite{biosnapnets}. BioSNAP also includes a small number of features for a selection of disease, side-effect and gene entities. This includes disease class information, pre-extracted network motifs, text-based synopses and structural pathway measures. Bio2RDF was an earlier project based on integrating disparate biological datastores~\cite{belleau2008bio2rdf}, using technologies from the Semantic Web stack. It incorporates data from 35 resources and the total Bio2RDF graph is over 10B triples. It is not explicitly focused towards the drug discovery domain and does not curate the resources in any way. In a similar vein, the Network Data Exchange (NDEx) is a repository for storing user submitted biological graphs, many of which are pertinent for drug discovery~\cite{pratt2015ndex}. BioGrakn is a biomedical knowledge project released as part of the GRAKN.AI graph analytics platform~\cite{messina2017biograkn}, including a collection of graphs, with the two most relevant being the precision medicine and disease focused graphs. 

\subsubsection{Graph Construction Resources} There are a growing number of resources designed to enable simpler construction of biomedical KGs, which include parsers for common resources, various filtering options and support for outputting into common formats. One such resource is Phenotype Knowledge Translator (PheKnowLator) which aims to build a framework for easier biomedical knowledge graph construction~\cite{callahan2020framework} and is available for public download\footnote{\url{https://github.com/callahantiff/PheKnowLator}}. Additionally, the Biological Expression Language (BEL), a domain-specific language for representing causal, correlative, and associative relationships between biological entities, is one way to help integrate disparate data sources~\cite{slater2014}. The Bio2BEL framework uses the PyBEL software ecosystem~\cite{hoyt2017} to programmatically convert biological data sources into BEL then to semantically align and merge them at scale. Each converter is loaded as a plugin such that the 70+ pre-defined converters can be used or new ones implemented, for example, to leverage data that is internal to the user's organization. Bio2BEL allows for \textit{ad hoc} KG assembly and is an alternative to static KG sources~\cite{ali2019biokeen}.
\section{Case Studies}
\label{sec:case_study}

In this section we highlight case studies from the literature, detailed in Table \ref{tab:case-studies}, where KGs have been successfully exploited in the drug discovery domain. We choose one approach from each of the areas of Polypharmacy, Drug-Target Interaction and Gene-Disease Prioritisation to highlight how KGs are being used across a range of tasks. We detail the successes, as well as analysing areas for further improvement.

\begin{table}[ht!] %
    \centering
    \resizebox{\textwidth}{!}{
    \begin{tabular}{p{0.18\textwidth} C{0.2\textwidth}  C{0.3\textwidth}  C{0.15\textwidth} G{0.1\textwidth} G{0.13\textwidth} G{0.1\textwidth}  G{0.15\textwidth} G{0.2\textwidth}} 
    \toprule
    \multicolumn{4}{c}{} & \multicolumn{5}{c}{\textbf{Dataset Information}}\\
    \cline{5-9}
    \textbf{Approach}  & \textbf{Domain} & \textbf{Model} & \textbf{Prediction Task} & \textbf{Entites} & \textbf{Relations} & \textbf{Entity Types} & \textbf{Relation Types} & \textbf{Num Datasets in Graph}\T\B \\
    \midrule \midrule
    Decagon~\cite{zitnik2018modeling} & Drug-Drug Interactions & Relational GCN with tensor factorisation decoder  &  Link Prediction & 19.6K & 5.3M & 2 & 964 & $\approx$7\T\B \\

    TriModel~\cite{mohamed2020discovering} & Drug-Target Interactions &  Tensor factorisation &  Link Prediction & 5K & 12K & 11 & 26 & 3\T\B \\

    Rosalind~\cite{paliwal2020preclinical} & Disease-Gene Prioritisation &  Tensor factorisation &  Link Prediction & 319K & 2.6M & 5 & 11 & $\approx$15\T\B \\

    \bottomrule
    \end{tabular}}
    \caption{An overview of drug discovery related approaches in the literature employing the use of KGs.}
    \label{tab:case-studies}
    \vskip -30pt
\end{table}

\subsection{Polypharmacy Prediction} 

The problem of adverse side effects that arises through the use of Polypharmacy (the use of more than one drug simultaneously to treat one or multiple conditions) has been modelled through the use of a KG and a novel GNN-based model entitled Decagon~\cite{zitnik2018modeling}.

\emph{Graph Composition.} The KG constructed was actually bipartite, containing only drug (over 900 unique entities) and protein (over 19K unique entities) entities~\cite{zitnik2018modeling}. These are linked through 964 unique edge types between drug-drug pairs, representing the various types of adverse side effects and a single edge type used to represent drug-protein and protein-protein interactions. Compared to the existing public KGs (Section \ref{sec:exisiting_kgs}), this graph places a lot of its complexity in the relation types, which has the potential drawback of limiting the amount of each seen during model training. Also unlike the existing public KGs, the graph is limited to just two entity types, suggesting disease or pathway information was deemed unimportant.

\emph{Underlying Datasets.} Data was extracted from protein centric databases like BioGRID~\cite{oughtred2019biogrid}, STRING~\cite{szklarczyk2019string}, STITCH~\cite{szklarczyk2016stitch}, as well as drug centric resources like SIDER~\cite{kuhn2016sider}, OFFSIDES, and TWOSIDES~\cite{tatonetti2012data}-- with much of the processed data being available in the BioSNAP project. Additionally, the graph is enriched with features on only the drug vertices containing descriptive single drug side-effect information.

\emph{Model.} The model encoder, similar to a Relational Graph Convolutional Network (R-GCN)~\cite{schlichtkrull2018modeling}, uses a separate parameter matrix for each edge type to learn relational aware vertex level embeddings. These embeddings are then input into a tensor factorisation-based decoder to directly predict potential negative drug-drug interactions via link prediction. The presented results show that compared with non-graph specific and homogeneous models, Decagon is better able to predict existing, and even propose novel, drug-drug interactions.

\subsection{Drug-Target Interaction Prediction}

Work has explored the task of Drug-Target Interaction prediction using a KG, containing existing protein and drug compound interactions, and a model entitled TriModel~\cite{mohamed2020discovering}

\emph{Graph Composition.} The graph is composed of 11 different entity types including genes and drugs, linked via 26 different relationship types. However, a precise discussion of why the chosen relation and entity types included in the graph is missing in the paper. Although the approach is focused on finding links between genes and drugs, only a single relation types is used to capture all possible interactions between these two entities.

\emph{Underlying Datasets.} The graph was constructed from KEGG, UniProt and DrugBank, with other known drug-target interaction resources such as Yamanishi08~\cite{yamanishi2008prediction} being used for evaluation. As such, the graph draws on fewer datasets than is typically seen in the existing public resources.

\emph{Model.} The authors propose that DTI can be formulated as a link-prediction task and introduce a model entitled TriModel in order to accomplish this. Similar to other Knowledge Graph Embedding (KGE) approaches~\cite{wang2017knowledge}, TriModel learns an embedding for all entities and relations in the graph by optimising the parameters such that true triplets are more accurately predicted over randomly sampled negatives. Over various traditional and non-relational graph methods, TriModel demonstrates superior performance, perhaps highlighting the importance of complex multi-relational information in generating accurate predictions.

\subsection{Gene-Disease Prioritisation}

The task of gene prioritisation (detailed in Section \ref{ssec:drug_discovery_intro}), has been addressed via the use of a KG~\cite{paliwal2020preclinical}. The overall approach, entitled Rosalind, details the construction of a knowledge graph and the choosing of a suitable model with which to make predictions. The work proposes that the disease target identification problem can be modelled as a link prediction task where the prediction of an edge between a disease and a gene entity would indicate possible association between the two. 

\emph{Graph Composition.} The Rosalind KG comprises five entity type (genes, compounds, diseases, bio-processes and pathways) linked via eleven relation types. As such, Rosalind most closely resembles existing KGs like Hetionet and BioKG in structure. Of note is that it captures some of the subtlety around disease-gene prioritisation, as ideally the model would predict which genes have some causal effect on the disease, not just an association. In Rosalind, they use two different types of edge between disease and gene entities -- one indicating association and the other therapeutic links (a drug exists targeting the gene to help alleviate disease). However, to date, the authors have not released the KG, making reproducibility challenging.

\emph{Underlying Datasets.} The Rosalind KG is constructed from many of the datasets detailed in this review. For example, the graph incorporates disease information from resources like DisGeNET, OMIM, and GWAS Catalog, interaction information from BioGRID, pathway information from Reactome and compound information from ChEMBL. 

\emph{Model.} The model chosen for the work is the ComplEx tensor factorisation approach~\cite{trouillon2016complex}. The evaluation of the approach demonstrates that it outperforms competing methods, including OpenTarget~\cite{carvalho2019open}, by as much as an extra 20\% of recall when predicting potential gene-disease relationships over 198 diseases. Model performance is evaluated only on this therapeutic edge type. Additionally, results are presented on a time-slices graph, where the model is trained on historical data and predictions are made on future edges. This is attempting to replicate the task we would ideally want performed - using the currently available knowledge to predict currently unknown information, in this case, unknown relationships between genes and diseases.

\subsection{Evaluative Summary}

These case studies have highlighted the considerable potential and successes of KGs aiding in a diverse set of drug discovery tasks. However, there are still areas for improvement regarding aspects of underlying data use and graph composition. Thus, we make the following observations:

\begin{itemize}
    \item \emph{Composition.} The composition of the KGs in these studies varies dramatically regarding entity and relation type quantities. This suggests there is not yet a consensus on the optimal way to compose drug discovery KGs for use in ML pipelines.
    \item \emph{Dataset Useage.} These studies use many of the datasets covered in this review to build their KGs. However there is still variety in where common relationship types are extracted from and usually no justification as to why a certain source was chosen.
    \item \emph{One graph to rule them all?} It is striking that none of the approaches utilise an existing KG. Instead, custom task-specific graphs are still typically created, perhaps highlighting the challenge in creating a single KG to address all possible tasks within drug discovery.
\end{itemize}
\section{Future Challenges \& Key Issues}
\label{sec:future}

Whilst there has been signifiant progress made in the field, there are still numerous open challenges and issues to be addressed. In this section, we detail major areas still needing improvement, which could help produce better drug discovery KGs. Here, we build upon many of the challenges of working with drug discovery data we established in Section \ref{ssec:graph-ml-kg} and the issues with pre-existing KGs in Section \ref{ssec:analysis-kg}.

\textbf{Graph Composition.} Constructing a useful KG for use in the drug discovery domain is still a challenging problem, especially when performed by non-domain experts. Many choices must be made when transforming a data source into a graph, especially if it is not relational by nature. Here, there is however great scope for interdisciplinary collaborations between domain scientists and KG and machine learning researchers. Additionally, we would like to see more high quality pre-constructed KGs, designed and validated by domain scientists, be made available for use by researchers. Further, creating graph construction toolkits, in which source datasets can be parsed in a unified and reproducible manner, would enable simpler creation of bespoke KGs.

\textbf{Data Value.} The availability of massive datasets has been partially credited with enabling the success of recent neural network models in areas such as computer vision~\cite{deng2009imagenet}. It might be tempting then to incorporate as much data as possible into drug discovery KG. However, much work still needs to be done in assessing the benefit of incorporating different data modalities. The consideration of value can also be extended to a financial view point: data collection, storage and processing can be expensive, especially if larger datasets do not improve performance in the task of interest. Another questions is whether a single \emph{super} graph should be created, which attempts to capture all knowledge around drug discovery, or whether smaller, more task specific, projections enable better predictions overall.

\textbf{Better Metadata.} As highlighted throughout the review, many of the core data resources are typically updated and refined at frequent intervals. However, many of the pre-existing KGs do not capture exactly which version of a certain resources was used during its construction. Storing this information might allow for better reproducibility, as well as measuring any change in predictive performance as the underlying knowledge is updated over time. Improved metadata could also capture if the relationship was taken from an expert curated, or automated pipeline. Additionally, graphs could provide common alternative identifiers (for example including both Entrez Gene and Ensembl identifiers for gene entities) as properties to enable easier incorporation of additional resources into the graph.

\textbf{Incorporation of Features.} Typically many existing KGs are provided as little more than edge lists, with models trying to make predictions using this relational information alone. Throughout this review, we have attempted to highlight where data resources may be used to add additional features for entities and relations. However, it is easier to imagine suitable features for certain entities (proteins and chemicals for example, where structural information could be incorporated) than others. Additionally, any potential benefits of incorporating these extra features would need to be assessed fairly. Nevertheless, we feel that there is scope for the incorporation of features to enable graph-specific neural models to be better exploited in the domain, with some recent promising work being demonstrated in the literature~\cite{zheng2020pharmkg}.

\textbf{Addressing Bias.} Many biases will be present in a drug discovery KG and any model being trained upon it, may have its predictive performance skewed away from under-represented, but potentially crucial relationships. Even manually curated resources may suffer incur bias from the person performing the curation. Practitioners should be aware of these issues and steps could be taken to mitigate them by, for example, reweighing the model training process. Additionally, users could consider removing over represented entities if they are confident that they are not required in the area of study. The lack of true negative samples in many graphs also means that the negative sampling strategy employed can bias the results. Recent inclusion of true negative samples in a benchmark graph~\cite{breit2020openbiolink} is encouraging, however where they are not possible to collect, more domain-aware sampling strategies should be investigated. 

\textbf{Fair Evaluation.} Due to the combinatorial ingestion process used to construct KGs, it is common for edges to be duplicated if the relationship is captured in more than one underling source. This can cause obvious issues when it comes to creating train/test splits for evaluation if the issues are not considered. Further, the presence of trivial inverse relationships, many of which may be present, can also skew performance metrics~\cite{toutanova2015observed}. It may also be more useful to assess model performance on more \emph{biologically meaningful} data splits, for example by splitting on disease or protein family. It could help move the field forward if meaningful splits for key tasks within drug discovery could be created by experts and made available for public use. 

\textbf{Uncertainty.} So much of the data represented in a biological KG is uncertain, either due to the nature of the experiment that generated it, or because it has been automatically mined from the literature. Yet this uncertainty is rarely represented in the graph itself, perhaps leading to a false sense of trust being created by the presence of certain relationships. We feel that more should be done to incorporate any uncertainty directly inside the KG. This could allow methods to directly learn from this information, thus creating better and more robust predictions.

\textbf{Reproducibility.} As in many areas of machine learning~\cite{dacrema2021troubling, errica2019fair, lipton2018troubling}, reproducibility of results is still a major issue in the KG field~\cite{ali2020bringing}. It is common for many papers to publish results without also providing the exact graph constructed to generate them. We believe further improvements in this area are essential for continued development in the field.

\section{Conclusion}
\label{sec:conclusion}

The use of KGs, combined with machine learning techniques, has the potential to help address key challenges in the field of drug discovery, with promising early applications already being demonstrated in the tasks of drug repositioning, drug-drug interactions and gene prioritisation. In this review we have presented an overview of the various key related datasets which could provide some of the fundamental building blocks for a hypothetical drug discovery KG. The review has also detailed and evaluated the range of pre-existing public KGs in the drug discovery domain. Additionally, we have highlighted the many pitfalls and challenges of working with drug discovery-based data and signposted key issues practitioners should consider when choosing suitable sources. 

Our hope is that this review of suitable data sources, combined with recent works evaluating graph-specific machine learning models in the context of drug discovery~\cite{gaudelet2020utilising}, can help guide researchers from across the KG mining and machine learning fields in applying state-of-the-art techniques in the field. Overall, we hope this review can serve as a catalyst in making the drug discovery domain more accessible, sparking new thought and innovation, whilst allowing researchers to more easily address key tasks within the domain, ultimately helping to improve and extend human life through new medicines.

\section*{Acknowledgement}

We would like to thank Manasa Ramakrishna, Ufuk Kirik, Benedek Rozemberczki, Natalie Kurbatova, Elizaveta Semenova and Claus Bendtsen for help and feedback throughout the preparation of this manuscript. Stephen Bonner is a fellow of the AstraZeneca postdoctoral program.

\bibliographystyle{plain}
\bibliography{RPbib}

\begin{thebibliography}{100}

\bibitem{ali2020bringing}
Mehdi Ali, Max Berrendorf, Charles~Tapley Hoyt, Laurent Vermue, Mikhail Galkin,
  Sahand Sharifzadeh, Asja Fischer, Volker Tresp, and Jens Lehmann.
\newblock Bringing light into the dark: A large-scale evaluation of knowledge
  graph embedding models under a unified framework.
\newblock {\em arXiv preprint arXiv:2006.13365}, 2020.

\bibitem{ali2019biokeen}
Mehdi Ali, Charles~Tapley Hoyt, Daniel Domingo-Fern{\'a}ndez, Jens Lehmann, and
  Hajira Jabeen.
\newblock Biokeen: a library for learning and evaluating biological knowledge
  graph embeddings.
\newblock {\em Bioinformatics}, 35(18):3538--3540, 2019.

\bibitem{antoniou2004semantic}
Grigoris Antoniou and Frank Van~Harmelen.
\newblock {\em A semantic web primer}.
\newblock MIT press, 2004.

\bibitem{apweiler2004uniprot}
Rolf Apweiler, Amos Bairoch, Cathy~H Wu, Winona~C Barker, Brigitte Boeckmann,
  Serenella Ferro, Elisabeth Gasteiger, Hongzhan Huang, Rodrigo Lopez, Michele
  Magrane, et~al.
\newblock Uniprot: the universal protein knowledgebase.
\newblock {\em Nucleic acids research}, 32(suppl\_1):D115--D119, 2004.

\bibitem{bagherian2020machine}
Maryam Bagherian, Elyas Sabeti, Kai Wang, Maureen~A Sartor, Zaneta
  Nikolovska-Coleska, and Kayvan Najarian.
\newblock Machine learning approaches and databases for prediction of
  drug--target interaction: a survey paper.
\newblock {\em Briefings in bioinformatics}, 2020.

\bibitem{barabasi2011network}
Albert-L{\'a}szl{\'o} Barab{\'a}si, Natali Gulbahce, and Joseph Loscalzo.
\newblock Network medicine: a network-based approach to human disease.
\newblock {\em Nature reviews genetics}, 12(1):56--68, 2011.

\bibitem{barabasi2004network}
Albert-Laszlo Barabasi and Zoltan~N Oltvai.
\newblock Network biology: understanding the cell's functional organization.
\newblock {\em Nature reviews genetics}, 5(2):101--113, 2004.

\bibitem{belleau2008bio2rdf}
Fran{\c{c}}ois Belleau, Marc-Alexandre Nolin, Nicole Tourigny, Philippe
  Rigault, and Jean Morissette.
\newblock Bio2rdf: towards a mashup to build bioinformatics knowledge systems.
\newblock {\em Journal of biomedical informatics}, 41(5):706--716, 2008.

\bibitem{bengio2013representation}
Yoshua Bengio, Aaron Courville, and Pascal Vincent.
\newblock Representation learning: A review and new perspectives.
\newblock {\em IEEE transactions on pattern analysis and machine intelligence},
  35(8):1798--1828, 2013.

\bibitem{berrendorf2020ambiguity}
Max Berrendorf, Evgeniy Faerman, Laurent Vermue, and Volker Tresp.
\newblock On the ambiguity of rank-based evaluation of entity alignment or link
  prediction methods.
\newblock {\em arXiv preprint arXiv:2002.06914}, 2020.

\bibitem{bettencourt2020exploring}
Joao~H Bettencourt-Silva, Natasha Mulligan, Charles Jochim, Nagesh Yadav,
  Walter Sedlazek, Vanessa Lopez, and Martin Gleize.
\newblock Exploring the social drivers of health during a pandemic: Leveraging
  knowledge graphs and population trends in covid-19.
\newblock {\em Studies in Health Technology and Informatics}, 275:6--11, 2020.

\bibitem{blanco2020new}
Maria-Jesus Blanco and Kevin~M Gardinier.
\newblock New chemical modalities and strategic thinking in early drug
  discovery, 2020.

\bibitem{breit2020openbiolink}
Anna Breit, Simon Ott, Asan Agibetov, and Matthias Samwald.
\newblock Openbiolink: A benchmarking framework for large-scale biomedical link
  prediction.
\newblock {\em Bioinformatics}, 2020.

\bibitem{brown2017standard}
Adam~S Brown and Chirag~J Patel.
\newblock A standard database for drug repositioning.
\newblock {\em Scientific data}, 4(1):1--7, 2017.

\bibitem{brown2015gene}
Garth~R Brown, Vichet Hem, Kenneth~S Katz, Michael Ovetsky, Craig Wallin, Olga
  Ermolaeva, Igor Tolstoy, Tatiana Tatusova, Kim~D Pruitt, Donna~R Maglott,
  et~al.
\newblock Gene: a gene-centered information resource at ncbi.
\newblock {\em Nucleic acids research}, 43(D1):D36--D42, 2015.

\bibitem{buniello2019nhgri}
Annalisa Buniello, Jacqueline A~L MacArthur, Maria Cerezo, Laura~W Harris,
  James Hayhurst, Cinzia Malangone, Aoife McMahon, Joannella Morales, Edward
  Mountjoy, Elliot Sollis, et~al.
\newblock The nhgri-ebi gwas catalog of published genome-wide association
  studies, targeted arrays and summary statistics 2019.
\newblock {\em Nucleic acids research}, 47(D1):D1005--D1012, 2019.

\bibitem{callahan2020framework}
Tiffany~J Callahan, Ignacio~J Tripodi, Lawrence~E Hunter, and William~A
  Baumgartner.
\newblock A framework for automated construction of heterogeneous large-scale
  biomedical knowledge graphs.
\newblock {\em bioRxiv}, 2020.

\bibitem{callahan2020knowledge}
Tiffany~J Callahan, Ignacio~J Tripodi, Harrison Pielke-Lombardo, and Lawrence~E
  Hunter.
\newblock Knowledge-based biomedical data science.
\newblock {\em Annual Review of Biomedical Data Science}, 3, 2020.

\bibitem{carvalho2019open}
Denise Carvalho-Silva, Andrea Pierleoni, Miguel Pignatelli, ChuangKee Ong, Luca
  Fumis, Nikiforos Karamanis, Miguel Carmona, Adam Faulconbridge, Andrew
  Hercules, Elaine McAuley, et~al.
\newblock Open targets platform: new developments and updates two years on.
\newblock {\em Nucleic acids research}, 47(D1):D1056--D1065, 2019.

\bibitem{celebi2019evaluation}
Remzi Celebi, Huseyin Uyar, Erkan Yasar, Ozgur Gumus, Oguz Dikenelli, and
  Michel Dumontier.
\newblock Evaluation of knowledge graph embedding approaches for drug-drug
  interaction prediction in realistic settings.
\newblock {\em BMC bioinformatics}, 20(1):1--14, 2019.

\bibitem{cernile2020network}
George Cernile, Trevor Heritage, Neil~J Sebire, Ben Gordon, Taralyn Schwering,
  Shana Kazemlou, and Yulia Borecki.
\newblock Network graph representation of covid-19 scientific publications to
  aid knowledge discovery.
\newblock {\em BMJ Health \& Care Informatics}, 28(1), 2020.

\bibitem{chen2018machine}
Ruolan Chen, Xiangrong Liu, Shuting Jin, Jiawei Lin, and Juan Liu.
\newblock Machine learning for drug-target interaction prediction.
\newblock {\em Molecules}, 23(9):2208, 2018.

\bibitem{chen2001bindingdb}
Xi~Chen, Ming Liu, and Michael~K Gilson.
\newblock Bindingdb: a web-accessible molecular recognition database.
\newblock {\em Combinatorial chemistry \& high throughput screening},
  4(8):719--725, 2001.

\bibitem{chen2002ttd}
Xin Chen, Zhi~Liang Ji, and Yu~Zong Chen.
\newblock Ttd: therapeutic target database.
\newblock {\em Nucleic acids research}, 30(1):412--415, 2002.

\bibitem{choobdar2019assessment}
Sarvenaz Choobdar, Mehmet~E Ahsen, Jake Crawford, Mattia Tomasoni, Tao Fang,
  David Lamparter, Junyuan Lin, Benjamin Hescott, Xiaozhe Hu, Johnathan Mercer,
  et~al.
\newblock Assessment of network module identification across complex diseases.
\newblock {\em Nature methods}, 16(9):843--852, 2019.

\bibitem{chou2016}
Chih~Hung Chou, Nai~Wen Chang, Sirjana Shrestha, Sheng~Da Hsu, Yu~Ling Lin,
  Wei~Hsiang Lee, Chi~Dung Yang, Hsiao~Chin Hong, Ting~Yen Wei, Siang~Jyun Tu,
  Tzi~Ren Tsai, Shu~Yi Ho, Ting~Yan Jian, Hsin~Yi Wu, Pin~Rong Chen, Nai~Chieh
  Lin, Hsin~Tzu Huang, Tzu~Ling Yang, Chung~Yuan Pai, Chun~San Tai, Wen~Liang
  Chen, Chia~Yen Huang, Chun~Chi Liu, Shun~Long Weng, Kuang~Wen Liao, Wen~Lian
  Hsu, and Hsien~Da Huang.
\newblock {miRTarBase 2016: Updates to the experimentally validated
  miRNA-target interactions database}.
\newblock {\em Nucleic Acids Research}, 44(D1):D239--D247, 2016.

\bibitem{gene2004gene}
Gene~Ontology Consortium.
\newblock The gene ontology (go) database and informatics resource.
\newblock {\em Nucleic acids research}, 32(suppl\_1):D258--D261, 2004.

\bibitem{gene2008gene}
Gene~Ontology Consortium.
\newblock The gene ontology project in 2008.
\newblock {\em Nucleic acids research}, 36(suppl\_1):D440--D444, 2008.

\bibitem{cook2014lessons}
David Cook, Dearg Brown, Robert Alexander, Ruth March, Paul Morgan, Gemma
  Satterthwaite, and Menelas~N Pangalos.
\newblock Lessons learned from the fate of astrazeneca's drug pipeline: a
  five-dimensional framework.
\newblock {\em Nature reviews Drug discovery}, 13(6):419--431, 2014.

\bibitem{dacrema2021troubling}
Maurizio~Ferrari Dacrema, Simone Boglio, Paolo Cremonesi, and Dietmar Jannach.
\newblock A troubling analysis of reproducibility and progress in recommender
  systems research.
\newblock {\em ACM Transactions on Information Systems (TOIS)}, 39(2):1--49,
  2021.

\bibitem{davis2019comparative}
Allan~Peter Davis, Cynthia~J Grondin, Robin~J Johnson, Daniela Sciaky, Roy
  McMorran, Jolene Wiegers, Thomas~C Wiegers, and Carolyn~J Mattingly.
\newblock The comparative toxicogenomics database: update 2019.
\newblock {\em Nucleic acids research}, 47(D1):D948--D954, 2019.

\bibitem{deng2009imagenet}
Jia Deng, Wei Dong, Richard Socher, Li-Jia Li, Kai Li, and Li~Fei-Fei.
\newblock Imagenet: A large-scale hierarchical image database.
\newblock In {\em 2009 IEEE conference on computer vision and pattern
  recognition}, pages 248--255. Ieee, 2009.

\bibitem{domingo2020covid}
Daniel Domingo-Fernandez, Shounak Baksi, Bruce Schultz, Yojana Gadiya, Reagon
  Karki, Tamara Raschka, Christian Ebeling, Martin Hofmann-Apitius, et~al.
\newblock Covid-19 knowledge graph: a computable, multi-modal, cause-and-effect
  knowledge model of covid-19 pathophysiology.
\newblock {\em Bioinformatics}, 09 2020.

\bibitem{durinx2016identifying}
Christine Durinx, Jo~McEntyre, Ron Appel, Rolf Apweiler, Mary Barlow, Niklas
  Blomberg, Chuck Cook, Elisabeth Gasteiger, Jee-Hyub Kim, Rodrigo Lopez,
  et~al.
\newblock Identifying elixir core data resources.
\newblock {\em F1000Research}, 5, 2016.

\bibitem{en2001ontologies}
Ste en~Schulze-Kremer.
\newblock Ontologies for molecular biology.
\newblock {\em Computer and Information Science}, 6(21), 2001.

\bibitem{errica2019fair}
Federico Errica, Marco Podda, Davide Bacciu, and Alessio Micheli.
\newblock A fair comparison of graph neural networks for graph classification.
\newblock {\em arXiv preprint arXiv:1912.09893}, 2019.

\bibitem{franccois2020dictionary}
Liesbeth Fran{\c{c}}ois, Jonathan van Eyll, and Patrice Godard.
\newblock Dictionary of disease ontologies (dodo): a graph database to
  facilitate access and interaction with disease and phenotype ontologies.
\newblock {\em F1000Research}, 9(942):942, 2020.

\bibitem{gaudelet2020utilising}
Thomas Gaudelet, Ben Day, Arian~R Jamasb, Jyothish Soman, Cristian Regep,
  Gertrude Liu, Jeremy B~R Hayter, Richard Vickers, Charles Roberts, Jian Tang,
  David Roblin, Tom~L Blundell, Michael~M Bronstein, and Jake~P Taylor-King.
\newblock {Utilizing graph machine learning within drug discovery and
  development}.
\newblock {\em Briefings in Bioinformatics}, 05 2021.

\bibitem{ghoussaini2020open}
Maya Ghoussaini, Edward Mountjoy, Miguel Carmona, Gareth Peat, Ellen~M Schmidt,
  Andrew Hercules, Luca Fumis, Alfredo Miranda, Denise Carvalho-Silva, Annalisa
  Buniello, et~al.
\newblock Open targets genetics: systematic identification of trait-associated
  genes using large-scale genetics and functional genomics.
\newblock {\em Nucleic Acids Research}, 2020.

\bibitem{hamilton2020graph}
William~L Hamilton.
\newblock Graph representation learning.
\newblock {\em Synthesis Lectures on Artificial Intelligence and Machine
  Learning}, 14(3):1--159.

\bibitem{hamosh2000online}
Ada Hamosh, Alan~F Scott, Joanna Amberger, David Valle, and Victor~A McKusick.
\newblock Online mendelian inheritance in man (omim).
\newblock {\em Human mutation}, 15(1):57--61, 2000.

\bibitem{have2013graph}
Christian~Theil Have and Lars~Juhl Jensen.
\newblock Are graph databases ready for bioinformatics?
\newblock {\em Bioinformatics}, 29(24):3107, 2013.

\bibitem{hermjakob2004intact}
Henning Hermjakob, Luisa Montecchi-Palazzi, Chris Lewington, Sugath Mudali,
  Samuel Kerrien, Sandra Orchard, Martin Vingron, Bernd Roechert, Peter
  Roepstorff, Alfonso Valencia, et~al.
\newblock Intact: an open source molecular interaction database.
\newblock {\em Nucleic acids research}, 32(suppl\_1):D452--D455, 2004.

\bibitem{himmelstein2017systematic}
Daniel~Scott Himmelstein, Antoine Lizee, Christine Hessler, Leo Brueggeman,
  Sabrina~L Chen, Dexter Hadley, Ari Green, Pouya Khankhanian, and Sergio~E
  Baranzini.
\newblock Systematic integration of biomedical knowledge prioritizes drugs for
  repurposing.
\newblock {\em Elife}, 6:e26726, 2017.

\bibitem{hirohara2018convolutional}
Maya Hirohara, Yutaka Saito, Yuki Koda, Kengo Sato, and Yasubumi Sakakibara.
\newblock Convolutional neural network based on smiles representation of
  compounds for detecting chemical motif.
\newblock {\em BMC bioinformatics}, 19(19):83--94, 2018.

\bibitem{hogan2021knowledge}
Aidan Hogan, Eva Blomqvist, Michael Cochez, Claudia d’Amato, Gerard~De Melo,
  Claudio Gutierrez, Sabrina Kirrane, Jos{\'e} Emilio~Labra Gayo, Roberto
  Navigli, Sebastian Neumaier, et~al.
\newblock Knowledge graphs.
\newblock {\em ACM Computing Surveys (CSUR)}, 54(4):1--37, 2021.

\bibitem{hornbeck2015}
Peter~V. Hornbeck, Bin Zhang, Beth Murray, Jon~M. Kornhauser, Vaughan Latham,
  and Elzbieta Skrzypek.
\newblock {PhosphoSitePlus, 2014: Mutations, PTMs and recalibrations}.
\newblock {\em Nucleic Acids Research}, 43(D1):D512--D520, 2015.

\bibitem{hoyt2017}
Charles~Tapley Hoyt, Andrej Konotopez, Christian Ebeling, and Jonathan Wren.
\newblock {PyBEL: a computational framework for Biological Expression
  Language.}
\newblock {\em Bioinformatics (Oxford, England)}, 34(4):703--704, feb 2018.

\bibitem{hughes2011principles}
James~P Hughes, Stephen Rees, S~Barrett Kalindjian, and Karen~L Philpott.
\newblock Principles of early drug discovery.
\newblock {\em British journal of pharmacology}, 162(6):1239--1249, 2011.

\bibitem{hwang2019humannet}
Sohyun Hwang, Chan~Yeong Kim, Sunmo Yang, Eiru Kim, Traver Hart, Edward~M
  Marcotte, and Insuk Lee.
\newblock Humannet v2: human gene networks for disease research.
\newblock {\em Nucleic acids research}, 47(D1):D573--D580, 2019.

\bibitem{drkg2020}
Vassilis~N. Ioannidis, Xiang Song, Saurav Manchanda, Mufei Li, Xiaoqin Pan,
  Da~Zheng, Xia Ning, Xiangxiang Zeng, and George Karypis.
\newblock Drkg - drug repurposing knowledge graph for covid-19.
\newblock \url{https://github.com/gnn4dr/DRKG/}, 2020.

\bibitem{ioannidis2020few}
Vassilis~N Ioannidis, Da~Zheng, and George Karypis.
\newblock Few-shot link prediction via graph neural networks for covid-19
  drug-repurposing.
\newblock {\em arXiv preprint arXiv:2007.10261}, 2020.

\bibitem{jassal2020reactome}
Bijay Jassal, Lisa Matthews, Guilherme Viteri, Chuqiao Gong, Pascual Lorente,
  Antonio Fabregat, Konstantinos Sidiropoulos, Justin Cook, Marc Gillespie,
  Robin Haw, et~al.
\newblock The reactome pathway knowledgebase.
\newblock {\em Nucleic acids research}, 48(D1):D498--D503, 2020.

\bibitem{jupp2017oxo}
Simon Jupp, Thomas Liener, Sirarat Sarntivijai, Olga Vrousgou, Tony Burdett,
  and Helen~E Parkinson.
\newblock Oxo-a gravy of ontology mapping extracts.
\newblock In {\em ICBO}, 2017.

\bibitem{jupp2014ebi}
Simon Jupp, James Malone, Jerven Bolleman, Marco Brandizi, Mark Davies, Leyla
  Garcia, Anna Gaulton, Sebastien Gehant, Camille Laibe, Nicole Redaschi,
  et~al.
\newblock The ebi rdf platform: linked open data for the life sciences.
\newblock {\em Bioinformatics}, 30(9):1338--1339, 2014.

\bibitem{kanehisa2007kegg}
Minoru Kanehisa, Michihiro Araki, Susumu Goto, Masahiro Hattori, Mika Hirakawa,
  Masumi Itoh, Toshiaki Katayama, Shuichi Kawashima, Shujiro Okuda, Toshiaki
  Tokimatsu, et~al.
\newblock Kegg for linking genomes to life and the environment.
\newblock {\em Nucleic acids research}, 36(suppl\_1):D480--D484, 2007.

\bibitem{kanehisa2017kegg}
Minoru Kanehisa, Miho Furumichi, Mao Tanabe, Yoko Sato, and Kanae Morishima.
\newblock Kegg: new perspectives on genomes, pathways, diseases and drugs.
\newblock {\em Nucleic acids research}, 45(D1):D353--D361, 2017.

\bibitem{kanehisa2010kegg}
Minoru Kanehisa, Susumu Goto, Miho Furumichi, Mao Tanabe, and Mika Hirakawa.
\newblock Kegg for representation and analysis of molecular networks involving
  diseases and drugs.
\newblock {\em Nucleic acids research}, 38(suppl\_1):D355--D360, 2010.

\bibitem{kim2016pubchem}
Sunghwan Kim, Paul~A Thiessen, Evan~E Bolton, Jie Chen, Gang Fu, Asta
  Gindulyte, Lianyi Han, Jane He, Siqian He, Benjamin~A Shoemaker, et~al.
\newblock Pubchem substance and compound databases.
\newblock {\em Nucleic acids research}, 44(D1):D1202--D1213, 2016.

\bibitem{king2019drug}
Emily~A King, J~Wade Davis, and Jacob~F Degner.
\newblock Are drug targets with genetic support twice as likely to be approved?
  revised estimates of the impact of genetic support for drug mechanisms on the
  probability of drug approval.
\newblock {\em PLoS genetics}, 15(12):e1008489, 2019.

\bibitem{kohler2019expansion}
Sebastian K{\"o}hler, Leigh Carmody, Nicole Vasilevsky, Julius O~B Jacobsen,
  Daniel Danis, Jean-Philippe Gourdine, Michael Gargano, Nomi~L Harris, Nicolas
  Matentzoglu, Julie~A McMurry, et~al.
\newblock Expansion of the human phenotype ontology (hpo) knowledge base and
  resources.
\newblock {\em Nucleic acids research}, 47(D1):D1018--D1027, 2019.

\bibitem{koscielny2017open}
Gautier Koscielny, Peter An, Denise Carvalho-Silva, Jennifer~A Cham, Luca
  Fumis, Rippa Gasparyan, Samiul Hasan, Nikiforos Karamanis, Michael Maguire,
  Eliseo Papa, et~al.
\newblock Open targets: a platform for therapeutic target identification and
  validation.
\newblock {\em Nucleic acids research}, 45(D1):D985--D994, 2017.

\bibitem{kuhn2016sider}
Michael Kuhn, Ivica Letunic, Lars~Juhl Jensen, and Peer Bork.
\newblock The sider database of drugs and side effects.
\newblock {\em Nucleic acids research}, 44(D1):D1075--D1079, 2016.

\bibitem{lee2020heterogeneous}
Bohyun Lee, Shuo Zhang, Aleksandar Poleksic, and Lei Xie.
\newblock Heterogeneous multi-layered network model for omics data integration
  and analysis.
\newblock {\em Frontiers in Genetics}, 10:1381, 2020.

\bibitem{lee2020biobert}
Jinhyuk Lee, Wonjin Yoon, Sungdong Kim, Donghyeon Kim, Sunkyu Kim, Chan~Ho So,
  and Jaewoo Kang.
\newblock Biobert: a pre-trained biomedical language representation model for
  biomedical text mining.
\newblock {\em Bioinformatics}, 36(4):1234--1240, 2020.

\bibitem{licata2012mint}
Luana Licata, Leonardo Briganti, Daniele Peluso, Livia Perfetto, Marta
  Iannuccelli, Eugenia Galeota, Francesca Sacco, Anita Palma, Aurelio~Pio
  Nardozza, Elena Santonico, et~al.
\newblock Mint, the molecular interaction database: 2012 update.
\newblock {\em Nucleic acids research}, 40(D1):D857--D861, 2012.

\bibitem{lin2017drug}
Yu~Lin, Saurabh Mehta, Hande K{\"u}{\c{c}}{\"u}k-McGinty, John~Paul Turner,
  Dusica Vidovic, Michele Forlin, Amar Koleti, Dac-Trung Nguyen, Lars~Juhl
  Jensen, Rajarshi Guha, et~al.
\newblock Drug target ontology to classify and integrate drug discovery data.
\newblock {\em Journal of biomedical semantics}, 8(1):50, 2017.

\bibitem{lindsay2003target}
Mark~A Lindsay.
\newblock Target discovery.
\newblock {\em Nature Reviews Drug Discovery}, 2(10):831--838, 2003.

\bibitem{lipscomb2000medical}
Carolyn~E Lipscomb.
\newblock Medical subject headings (mesh).
\newblock {\em Bulletin of the Medical Library Association}, 88(3):265, 2000.

\bibitem{lipton2018troubling}
Zachary~C Lipton and Jacob Steinhardt.
\newblock Troubling trends in machine learning scholarship.
\newblock {\em arXiv preprint arXiv:1807.03341}, 2018.

\bibitem{Lopez-DelRio2019}
Angela {Lopez-Del Rio}, Alfons Nonell-Canals, David Vidal, and Alexandre
  Perera-Lluna.
\newblock {Evaluation of Cross-Validation Strategies in Sequence-Based Binding
  Prediction Using Deep Learning}.
\newblock {\em Journal of Chemical Information and Modeling}, 59(4):1645--1657,
  2019.

\bibitem{luo2020biomedical}
H~Luo, M~Li, M~Yang, FX~Wu, Y~Li, and J~Wang.
\newblock Biomedical data and computational models for drug repositioning: a
  comprehensive review.
\newblock {\em Briefings in Bioinformatics}, 2020.

\bibitem{maglott2005entrez}
Donna Maglott, Jim Ostell, Kim~D Pruitt, and Tatiana Tatusova.
\newblock Entrez gene: gene-centered information at ncbi.
\newblock {\em Nucleic acids research}, 33(suppl\_1):D54--D58, 2005.

\bibitem{malone2010modeling}
James Malone, Ele Holloway, Tomasz Adamusiak, Misha Kapushesky, Jie Zheng,
  Nikolay Kolesnikov, Anna Zhukova, Alvis Brazma, and Helen Parkinson.
\newblock Modeling sample variables with an experimental factor ontology.
\newblock {\em Bioinformatics}, 26(8):1112--1118, 2010.

\bibitem{masoudi2020drug}
Yosef Masoudi-Sobhanzadeh, Yadollah Omidi, Massoud Amanlou, and Ali
  Masoudi-Nejad.
\newblock Drug databases and their contributions to drug repurposing.
\newblock {\em Genomics}, 112(2):1087--1095, 2020.

\bibitem{mendez2019chembl}
David Mendez, Anna Gaulton, A~Patr{\'\i}cia Bento, Jon Chambers, Marleen
  De~Veij, Eloy F{\'e}lix, Mar{\'\i}a~Paula Magari{\~n}os, Juan~F Mosquera,
  Prudence Mutowo, Micha{\l} Nowotka, et~al.
\newblock Chembl: towards direct deposition of bioassay data.
\newblock {\em Nucleic acids research}, 47(D1):D930--D940, 2019.

\bibitem{messina2017biograkn}
Antonio Messina, Haikal Pribadi, Jo~Stichbury, Michelangelo Bucci, Szymon
  Klarman, and Alfonso Urso.
\newblock Biograkn: A knowledge graph-based semantic database for biomedical
  sciences.
\newblock In {\em Conference on Complex, Intelligent, and Software Intensive
  Systems}, pages 299--309. Springer, 2017.

\bibitem{mikolov2013distributed}
Tomas Mikolov, Ilya Sutskever, Kai Chen, Greg~S Corrado, and Jeff Dean.
\newblock Distributed representations of words and phrases and their
  compositionality.
\newblock {\em Advances in Neural Information Processing Systems},
  26:3111--3119, 2013.

\bibitem{mohamed2020biological}
Sameh~K Mohamed, Aayah Nounu, and V{\'\i}t Nov{\'a}{\v{c}}ek.
\newblock Biological applications of knowledge graph embedding models.
\newblock {\em Briefings in Bioinformatics}, 2020.

\bibitem{mohamed2020discovering}
Sameh~K Mohamed, V{\'\i}t Nov{\'a}{\v{c}}ek, and Aayah Nounu.
\newblock Discovering protein drug targets using knowledge graph embeddings.
\newblock {\em Bioinformatics}, 36(2):603--610, 2020.

\bibitem{morgan2018impact}
Paul Morgan, Dean~G Brown, Simon Lennard, Mark~J Anderton, J~Carl Barrett, Ulf
  Eriksson, Mark Fidock, Bengt Hamren, Anthony Johnson, Ruth~E March, et~al.
\newblock Impact of a five-dimensional framework on r\&d productivity at
  astrazeneca.
\newblock {\em Nature reviews Drug discovery}, 17(3):167, 2018.

\bibitem{mubeen2019}
Sarah Mubeen, Charles~Tapley Hoyt, Andr{\'{e}} Gem{\"{u}}nd, Martin
  Hofmann-Apitius, Holger Fr{\"{o}}hlich, and Daniel Domingo-Fern{\'{a}}ndez.
\newblock {The Impact of Pathway Database Choice on Statistical Enrichment
  Analysis and Predictive Modeling}.
\newblock {\em Frontiers in Genetics}, 10:654442, nov 2019.

\bibitem{mungall2017monarch}
Christopher~J Mungall, Julie~A McMurry, Sebastian K{\"o}hler, James~P Balhoff,
  Charles Borromeo, Matthew Brush, Seth Carbon, Tom Conlin, Nathan Dunn, Mark
  Engelstad, et~al.
\newblock The monarch initiative: an integrative data and analytic platform
  connecting phenotypes to genotypes across species.
\newblock {\em Nucleic acids research}, 45(D1):D712--D722, 2017.

\bibitem{nelson2019integrating}
Charlotte~A Nelson, Atul~J Butte, and Sergio~E Baranzini.
\newblock Integrating biomedical research and electronic health records to
  create knowledge-based biologically meaningful machine-readable embeddings.
\newblock {\em Nature communications}, 10(1):1--10, 2019.

\bibitem{nelson2015support}
Matthew~R Nelson, Hannah Tipney, Jeffery~L Painter, Judong Shen, Paola
  Nicoletti, Yufeng Shen, Aris Floratos, Pak~Chung Sham, Mulin~Jun Li, Junwen
  Wang, et~al.
\newblock The support of human genetic evidence for approved drug indications.
\newblock {\em Nature genetics}, 47(8):856--860, 2015.

\bibitem{neyshabur2013netal}
Behnam Neyshabur, Ahmadreza Khadem, Somaye Hashemifar, and Seyed~Shahriar Arab.
\newblock Netal: a new graph-based method for global alignment of
  protein--protein interaction networks.
\newblock {\em Bioinformatics}, 29(13):1654--1662, 2013.

\bibitem{nguyen2017pharos}
Dac-Trung Nguyen, Stephen Mathias, Cristian Bologa, Soren Brunak, Nicolas
  Fernandez, Anna Gaulton, Anne Hersey, Jayme Holmes, Lars~Juhl Jensen, Anneli
  Karlsson, et~al.
\newblock Pharos: Collating protein information to shed light on the druggable
  genome.
\newblock {\em Nucleic acids research}, 45(D1):D995--D1002, 2017.

\bibitem{oprea2018unexplored}
Tudor~I Oprea, Cristian~G Bologa, S{\o}ren Brunak, Allen Campbell, Gregory~N
  Gan, Anna Gaulton, Shawn~M Gomez, Rajarshi Guha, Anne Hersey, Jayme Holmes,
  et~al.
\newblock Unexplored therapeutic opportunities in the human genome.
\newblock {\em Nature reviews Drug discovery}, 17(5):317, 2018.

\bibitem{oughtred2019biogrid}
Rose Oughtred, Chris Stark, Bobby-Joe Breitkreutz, Jennifer Rust, Lorrie
  Boucher, Christie Chang, Nadine Kolas, Lara O’Donnell, Genie Leung,
  Rochelle McAdam, et~al.
\newblock The biogrid interaction database: 2019 update.
\newblock {\em Nucleic acids research}, 47(D1):D529--D541, 2019.

\bibitem{paliwal2020preclinical}
Saee Paliwal, Alex de~Giorgio, Daniel Neil, Jean-Baptiste Michel, and Alix~MB
  Lacoste.
\newblock Preclinical validation of therapeutic targets predicted by tensor
  factorization on heterogeneous graphs.
\newblock {\em Scientific reports}, 10(1):1--19, 2020.

\bibitem{percha2018global}
Bethany Percha and Russ~B Altman.
\newblock A global network of biomedical relationships derived from text.
\newblock {\em Bioinformatics}, 34(15):2614--2624, 2018.

\bibitem{pinero2015disgenet}
Janet Pi{\~n}ero, N{\'u}ria Queralt-Rosinach, Alex Bravo, Jordi Deu-Pons, Anna
  Bauer-Mehren, Martin Baron, Ferran Sanz, and Laura~I Furlong.
\newblock Disgenet: a discovery platform for the dynamical exploration of human
  diseases and their genes.
\newblock {\em Database}, 2015, 2015.

\bibitem{pinero2020disgenet}
Janet Pi{\~n}ero, Juan~Manuel Ram{\'\i}rez-Anguita, Josep Sa{\"u}ch-Pitarch,
  Francesco Ronzano, Emilio Centeno, Ferran Sanz, and Laura~I Furlong.
\newblock The disgenet knowledge platform for disease genomics: 2019 update.
\newblock {\em Nucleic acids research}, 48(D1):D845--D855, 2020.

\bibitem{pletscher2015diseases}
Sune Pletscher-Frankild, Albert Pallej{\`a}, Kalliopi Tsafou, Janos~X Binder,
  and Lars~Juhl Jensen.
\newblock Diseases: Text mining and data integration of disease--gene
  associations.
\newblock {\em Methods}, 74:83--89, 2015.

\bibitem{pratt2015ndex}
Dexter Pratt, Jing Chen, David Welker, Ricardo Rivas, Rudolf Pillich, Vladimir
  Rynkov, Keiichiro Ono, Carol Miello, Lyndon Hicks, Sandor Szalma, et~al.
\newblock Ndex, the network data exchange.
\newblock {\em Cell systems}, 1(4):302--305, 2015.

\bibitem{reese2020kg}
Justin~T Reese, Deepak Unni, Tiffany~J Callahan, Luca Cappelletti, Vida
  Ravanmehr, Seth Carbon, Kent~A Shefchek, Benjamin~M Good, James~P Balhoff,
  Tommaso Fontana, et~al.
\newblock Kg-covid-19: a framework to produce customized knowledge graphs for
  covid-19 response.
\newblock {\em Patterns}, page 100155, 2020.

\bibitem{rigden202027th}
Daniel~J Rigden and Xos{\'e}~M Fern{\'a}ndez.
\newblock The 27th annual nucleic acids research database issue and molecular
  biology database collection.
\newblock {\em Nucleic Acids Research}, 48(D1):D1--D8, 2020.

\bibitem{rives2021biological}
Alexander Rives, Joshua Meier, Tom Sercu, Siddharth Goyal, Zeming Lin, Jason
  Liu, Demi Guo, Myle Ott, C~Lawrence Zitnick, Jerry Ma, et~al.
\newblock Biological structure and function emerge from scaling unsupervised
  learning to 250 million protein sequences.
\newblock {\em Proceedings of the National Academy of Sciences}, 118(15), 2021.

\bibitem{robinson2008human}
Peter~N Robinson, Sebastian K{\"o}hler, Sebastian Bauer, Dominik Seelow, Denise
  Horn, and Stefan Mundlos.
\newblock The human phenotype ontology: a tool for annotating and analyzing
  human hereditary disease.
\newblock {\em The American Journal of Human Genetics}, 83(5):610--615, 2008.

\bibitem{rubin2008biomedical}
Daniel~L Rubin, Nigam~H Shah, and Natalya~F Noy.
\newblock Biomedical ontologies: a functional perspective.
\newblock {\em Briefings in bioinformatics}, 9(1):75--90, 2008.

\bibitem{santos2020}
Alberto Santos, Ana~R Cola{\c{c}}o, Annelaura~B Nielsen, Lili Niu, Philipp~E
  Geyer, Fabian Coscia, Nicolai J~Wewer Albrechtsen, Filip Mundt, Lars~Juhl
  Jensen, and Matthias Mann.
\newblock {Clinical Knowledge Graph Integrates Proteomics Data into Clinical
  Decision-Making}.
\newblock {\em bioRxiv}, page 2020.05.09.084897, jan 2020.

\bibitem{schlichtkrull2018modeling}
Michael Schlichtkrull, Thomas~N Kipf, Peter Bloem, Rianne Van Den~Berg, Ivan
  Titov, and Max Welling.
\newblock Modeling relational data with graph convolutional networks.
\newblock In {\em European Semantic Web Conference}, pages 593--607. Springer,
  2018.

\bibitem{schriml2019human}
Lynn~M Schriml, Elvira Mitraka, James Munro, Becky Tauber, Mike Schor, Lance
  Nickle, Victor Felix, Linda Jeng, Cynthia Bearer, Richard Lichenstein, et~al.
\newblock Human disease ontology 2018 update: classification, content and
  workflow expansion.
\newblock {\em Nucleic acids research}, 47(D1):D955--D962, 2019.

\bibitem{slater2014}
Ted Slater.
\newblock {Recent advances in modeling languages for pathway maps and
  computable biological networks}.
\newblock {\em Drug Discovery Today}, 19(2):193--198, feb 2014.

\bibitem{slenter2018wikipathways}
Denise~N Slenter, Martina Kutmon, Kristina Hanspers, Anders Riutta, Jacob
  Windsor, Nuno Nunes, Jonathan M{\'e}lius, Elisa Cirillo, Susan~L Coort,
  Daniela Digles, et~al.
\newblock Wikipathways: a multifaceted pathway database bridging metabolomics
  to other omics research.
\newblock {\em Nucleic acids research}, 46(D1):D661--D667, 2018.

\bibitem{sorger2011quantitative}
Peter~K Sorger, Sandra~RB Allerheiligen, Darrell~R Abernethy, Russ~B Altman,
  Kim~LR Brouwer, Andrea Califano, David~Z D’Argenio, Ravi Iyengar, William~J
  Jusko, Richard Lalonde, et~al.
\newblock Quantitative and systems pharmacology in the post-genomic era: new
  approaches to discovering drugs and understanding therapeutic mechanisms.
\newblock In {\em An NIH white paper by the QSP workshop group}, volume~48. NIH
  Bethesda Bethesda, MD, 2011.

\bibitem{stark2006biogrid}
Chris Stark, Bobby-Joe Breitkreutz, Teresa Reguly, Lorrie Boucher, Ashton
  Breitkreutz, and Mike Tyers.
\newblock Biogrid: a general repository for interaction datasets.
\newblock {\em Nucleic acids research}, 34(suppl\_1):D535--D539, 2006.

\bibitem{sweeney2019}
Blake~A Sweeney, Anton~I Petrov, Boris Burkov, Robert~D Finn, Alex Bateman,
  Maciej Szymanski, Wojciech~M Karlowski, Jan Gorodkin, Stefan~E Seemann,
  Jamie~J Cannone, Robin~R Gutell, Petra Fey, Siddhartha Basu, Simon Kay, Guy
  Cochrane, Kostantinos Billis, David Emmert, Steven~J Marygold, Rachael~P
  Huntley, Ruth~C Lovering, Adam Frankish, Patricia~P Chan, Todd~M Lowe,
  Elspeth Bruford, Ruth Seal, Jo~Vandesompele, Pieter-Jan Volders, Maria
  Paraskevopoulou, Lina Ma, Zhang Zhang, Sam Griffiths-Jones, Janusz~M
  Bujnicki, Pietro Boccaletto, Judith~A Blake, Carol~J Bult, Runsheng Chen,
  Yi~Zhao, Valerie Wood, Kim Rutherford, Elena Rivas, James Cole, Stanley J~F
  Laulederkind, Mary Shimoyama, Marc~E Gillespie, Marija Orlic-Milacic, Ioanna
  Kalvari, Eric Nawrocki, Stacia~R Engel, J~Michael Cherry, SILVA Team, Tanya~Z
  Berardini, Artemis Hatzigeorgiou, Dimitra Karagkouni, Kevin Howe, Paul Davis,
  Marcel Dinger, Shunmin He, Maki Yoshihama, Naoya Kenmochi, Peter~F Stadler,
  and Kelly~P Williams.
\newblock {RNAcentral: a hub of information for non-coding RNA sequences}.
\newblock {\em Nucleic Acids Research}, 47(D1):D1250--D1251, jan 2019.

\bibitem{szklarczyk2019string}
Damian Szklarczyk, Annika~L Gable, David Lyon, Alexander Junge, Stefan Wyder,
  Jaime Huerta-Cepas, Milan Simonovic, Nadezhda~T Doncheva, John~H Morris, Peer
  Bork, et~al.
\newblock String v11: protein--protein association networks with increased
  coverage, supporting functional discovery in genome-wide experimental
  datasets.
\newblock {\em Nucleic acids research}, 47(D1):D607--D613, 2019.

\bibitem{szklarczyk2016stitch}
Damian Szklarczyk, Alberto Santos, Christian von Mering, Lars~Juhl Jensen, Peer
  Bork, and Michael Kuhn.
\newblock Stitch 5: augmenting protein--chemical interaction networks with
  tissue and affinity data.
\newblock {\em Nucleic acids research}, 44(D1):D380--D384, 2016.

\bibitem{tanoli2020exploration}
Ziaurrehman Tanoli, Umair Seemab, Andreas Scherer, Krister Wennerberg, Jing
  Tang, and Markus V{\"a}h{\"a}-Koskela.
\newblock Exploration of databases and methods supporting drug repurposing: a
  comprehensive survey.
\newblock {\em Briefings in Bioinformatics}, 2020.

\bibitem{tatonetti2012data}
Nicholas~P Tatonetti, P~Ye Patrick, Roxana Daneshjou, and Russ~B Altman.
\newblock Data-driven prediction of drug effects and interactions.
\newblock {\em Science translational medicine}, 4(125):125ra31--125ra31, 2012.

\bibitem{terstappen2001silico}
Georg~C Terstappen and Angelo Reggiani.
\newblock In silico research in drug discovery.
\newblock {\em Trends in pharmacological sciences}, 22(1):23--26, 2001.

\bibitem{tirmizi2011mapping}
Syed~Hamid Tirmizi, Stuart Aitken, Dilvan~A Moreira, Chris Mungall, Juan
  Sequeda, Nigam~H Shah, and Daniel~P Miranker.
\newblock Mapping between the obo and owl ontology languages.
\newblock {\em Journal of biomedical semantics}, 2(S1):S3, 2011.

\bibitem{toutanova2015observed}
Kristina Toutanova and Danqi Chen.
\newblock Observed versus latent features for knowledge base and text
  inference.
\newblock In {\em Proceedings of the 3rd workshop on continuous vector space
  models and their compositionality}, pages 57--66, 2015.

\bibitem{trouillon2016complex}
Th{\'e}o Trouillon, Johannes Welbl, Sebastian Riedel, {\'E}ric Gaussier, and
  Guillaume Bouchard.
\newblock Complex embeddings for simple link prediction.
\newblock International Conference on Machine Learning (ICML), 2016.

\bibitem{turei2016omnipath}
D{\'e}nes T{\"u}rei, Tam{\'a}s Korcsm{\'a}ros, and Julio Saez-Rodriguez.
\newblock Omnipath: guidelines and gateway for literature-curated signaling
  pathway resources.
\newblock {\em Nature methods}, 13(12):966--967, 2016.

\bibitem{ursu2016drugcentral}
Oleg Ursu, Jayme Holmes, Jeffrey Knockel, Cristian~G Bologa, Jeremy~J Yang,
  Stephen~L Mathias, Stuart~J Nelson, and Tudor~I Oprea.
\newblock Drugcentral: online drug compendium.
\newblock {\em Nucleic acids research}, page gkw993, 2016.

\bibitem{vamathevan2019applications}
Jessica Vamathevan, Dominic Clark, Paul Czodrowski, Ian Dunham, Edgardo Ferran,
  George Lee, Bin Li, Anant Madabhushi, Parantu Shah, Michaela Spitzer, et~al.
\newblock Applications of machine learning in drug discovery and development.
\newblock {\em Nature Reviews Drug Discovery}, 18(6):463--477, 2019.

\bibitem{wagner2018dynamic}
John Wagner, Andrew~M Dahlem, Lynn~D Hudson, Sharon~F Terry, Russ~B Altman,
  C~Taylor Gilliland, Christopher DeFeo, and Christopher~P Austin.
\newblock A dynamic map for learning, communicating, navigating and improving
  therapeutic development.
\newblock {\em Nature Reviews Drug Discovery}, 17(2):150--150, 2018.

\bibitem{walsh2020biokg}
Brian Walsh, Sameh~K Mohamed, and V{\'\i}t Nov{\'a}{\v{c}}ek.
\newblock Biokg: A knowledge graph for relational learning on biological data.
\newblock In {\em Proceedings of the 29th ACM International Conference on
  Information \& Knowledge Management}, pages 3173--3180, 2020.

\bibitem{wang2019dgl}
Minjie Wang, Da~Zheng, Zihao Ye, Quan Gan, Mufei Li, Xiang Song, Jinjing Zhou,
  Chao Ma, Lingfan Yu, Yu~Gai, Tianjun Xiao, Tong He, George Karypis, Jinyang
  Li, and Zheng Zhang.
\newblock Deep graph library: A graph-centric, highly-performant package for
  graph neural networks.
\newblock {\em arXiv preprint arXiv:1909.01315}, 2019.

\bibitem{wang2017knowledge}
Quan Wang, Zhendong Mao, Bin Wang, and Li~Guo.
\newblock Knowledge graph embedding: A survey of approaches and applications.
\newblock {\em IEEE Transactions on Knowledge and Data Engineering},
  29(12):2724--2743, 2017.

\bibitem{whirl2012pharmacogenomics}
Michelle Whirl-Carrillo, Ellen~M McDonagh, JM~Hebert, Li~Gong, K~Sangkuhl,
  CF~Thorn, Russ~B Altman, and Teri~E Klein.
\newblock Pharmacogenomics knowledge for personalized medicine.
\newblock {\em Clinical Pharmacology \& Therapeutics}, 92(4):414--417, 2012.

\bibitem{wise2020covid}
Colby Wise, Miguel~Romero Calvo, Pariminder Bhatia, Vassilis Ioannidis, George
  Karypus, George Price, Xiang Song, Ryan Brand, and Ninad Kulkani.
\newblock Covid-19 knowledge graph: Accelerating information retrieval and
  discovery for scientific literature.
\newblock In {\em Proceedings of Knowledgeable NLP: the First Workshop on
  Integrating Structured Knowledge and Neural Networks for NLP}, pages 1--10,
  2020.

\bibitem{wishart2018drugbank}
David~S Wishart, Yannick~D Feunang, An~C Guo, Elvis~J Lo, Ana Marcu, Jason~R
  Grant, Tanvir Sajed, Daniel Johnson, Carin Li, Zinat Sayeeda, et~al.
\newblock Drugbank 5.0: a major update to the drugbank database for 2018.
\newblock {\em Nucleic acids research}, 46(D1):D1074--D1082, 2018.

\bibitem{wishart2008drugbank}
David~S Wishart, Craig Knox, An~Chi Guo, Dean Cheng, Savita Shrivastava, Dan
  Tzur, Bijaya Gautam, and Murtaza Hassanali.
\newblock Drugbank: a knowledgebase for drugs, drug actions and drug targets.
\newblock {\em Nucleic acids research}, 36(suppl\_1):D901--D906, 2008.

\bibitem{yamanishi2008prediction}
Yoshihiro Yamanishi, Michihiro Araki, Alex Gutteridge, Wataru Honda, and Minoru
  Kanehisa.
\newblock Prediction of drug--target interaction networks from the integration
  of chemical and genomic spaces.
\newblock {\em Bioinformatics}, 24(13):i232--i240, 2008.

\bibitem{yates2020ensembl}
Andrew~D Yates, Premanand Achuthan, Wasiu Akanni, James Allen, Jamie Allen,
  Jorge Alvarez-Jarreta, M~Ridwan Amode, Irina~M Armean, Andrey~G Azov, Ruth
  Bennett, et~al.
\newblock Ensembl 2020.
\newblock {\em Nucleic acids research}, 48(D1):D682--D688, 2020.

\bibitem{zhang2019heterogeneous}
Chuxu Zhang, Dongjin Song, Chao Huang, Ananthram Swami, and Nitesh~V Chawla.
\newblock Heterogeneous graph neural network.
\newblock In {\em Proceedings of the 25th ACM SIGKDD International Conference
  on Knowledge Discovery \& Data Mining}, pages 793--803, 2019.

\bibitem{zhang2019nscaching}
Yongqi Zhang, Quanming Yao, Yingxia Shao, and Lei Chen.
\newblock Nscaching: simple and efficient negative sampling for knowledge graph
  embedding.
\newblock In {\em 2019 IEEE 35th International Conference on Data Engineering
  (ICDE)}, pages 614--625. IEEE, 2019.

\bibitem{zheng2020pharmkg}
Shuangjia Zheng, Jiahua Rao, Ying Song, Jixian Zhang, Xianglu Xiao, Evandro~Fei
  Fang, Yuedong Yang, and Zhangming Niu.
\newblock Pharmkg: a dedicated knowledge graph benchmark for bomedical data
  mining.
\newblock {\em Briefings in Bioinformatics}, 2020.

\bibitem{zhu2020knowledge}
Yongjun Zhu, Chao Che, Bo~Jin, Ningrui Zhang, Chang Su, and Fei Wang.
\newblock Knowledge-driven drug repurposing using a comprehensive drug
  knowledge graph.
\newblock {\em Health Informatics Journal}, page 1460458220937101, 2020.

\bibitem{zhu2019drug}
Yongjun Zhu, Olivier Elemento, Jyotishman Pathak, and Fei Wang.
\newblock Drug knowledge bases and their applications in biomedical informatics
  research.
\newblock {\em Briefings in bioinformatics}, 20(4):1308--1321, 2019.

\bibitem{zitnik2018modeling}
Marinka Zitnik, Monica Agrawal, and Jure Leskovec.
\newblock Modeling polypharmacy side effects with graph convolutional networks.
\newblock {\em Bioinformatics}, 34(13):i457--i466, 2018.

\bibitem{biosnapnets}
Marinka Zitnik, Rok Sosic, and Jure Leskovec.
\newblock Biosnap datasets: Stanford biomedical network dataset collection.
\newblock \url{http://snap.stanford.edu/biodata}, aug 2018.

\end{thebibliography}

\end{document}